\documentclass[sigconf,10pt]{acmart}

\usepackage{appendix}
\usepackage{dblfloatfix}
\usepackage{color}
\usepackage{colortbl}
\usepackage{multirow}
\usepackage[english]{babel}
\usepackage[utf8]{inputenc}
\usepackage{amsmath}

\usepackage{enumitem}
\usepackage{graphicx}
\usepackage{booktabs, makecell, multirow}
\usepackage{subcaption}
\usepackage{float}
\usepackage{bbm}
\usepackage{balance}
\usepackage{xcolor}
\usepackage{algorithm2e}
\usepackage{algpseudocode}
\usepackage{soul}
\usepackage{listings}
\lstdefinestyle{thin}{
  basicstyle=\ttfamily\small,
  numbers=left,
  numberstyle=\tiny,
  numbersep=6pt,
  frame=single,
  breaklines=true,
  tabsize=2,
  columns=fullflexible,
  keepspaces=true,
  showstringspaces=false
}

\usepackage{tabularx,array}
\newcolumntype{L}[1]{>{\raggedright\arraybackslash}p{#1}}
\newcolumntype{Y}{>{\raggedright\arraybackslash}X}

\emergencystretch=\maxdimen
\definecolor{LightGrey}{rgb}{0.8,0.8,0.8}

\newcommand{\BCUnit}{{\em ng/$m^{3}$}}  

\AtBeginDocument{%
  \providecommand\BibTeX{{%
    \normalfont B\kern-0.5em{\scshape i\kern-0.25em b}\kern-0.8em\TeX}}}

\setcopyright{acmlicensed}
\copyrightyear{2025}
\acmYear{2025}
\acmDOI{XXXXXXX.XXXXXXX}

\acmConference[MobiSys '26]{24th ACM International Conference on Mobile Systems, Applications, and Services}{XXX XX--XX, 2026}{XXX, XXX, XXX}

\graphicspath{{./images/}} 
\begin{document}

\title[Black Traces in Traffic]{Estimating Black Carbon Concentration from Urban Traffic Using Vision-Based Machine Learning}

\author{Camellia Zakaria}
\affiliation{
  \institution{University of Toronto}
  \country{Canada}
}
\email{camellia.zakaria@utoronto.ca}

\author{Aryan Sadeghi}
\affiliation{
  \institution{University of Toronto}
  \country{Canada}
}
\email{aryan.sadeghi@utoronto.ca}

\author{Weaam Jaafar}
\affiliation{
  \institution{University of Toronto}
  \country{Canada}
}
\email{weaam.jaafar@mail.utoronto.ca}

\author{Junshi Xu}
\affiliation{
  \institution{University of Hong Kong}
  \country{Hong Kong}
}
\email{junshixu@hku.hk}

\author{Alex Mariakakis}
\affiliation{
  \institution{University of Toronto}
  \country{Canada}
}
\email{alex.mariakakis@utoronto.ca}

\author{Marianne Hatzopoulou}
\affiliation{
  \institution{University of Toronto}
  \country{Canada}
}
\email{marianne.hatzopoulou@utoronto.ca}

\renewcommand{\shortauthors}{Zakaria et al.}
\begin{abstract}
Black carbon (BC) emissions in urban areas are primarily driven by traffic, with hotspots near major roads disproportionately affecting marginalized communities. Because BC monitoring is typically performed using costly and specialized instruments, there is little to no available data on BC concentration from local traffic sources that could help inform policy interventions targeting local factors. By contrast, traffic monitoring systems are widely deployed in cities around the world, highlighting the imbalance between what we know about traffic conditions and what we do not know about their environmental consequences. To bridge this gap, we propose a machine learning–driven system that extracts visual information from traffic video to capture vehicle behaviors and conditions. Combining these features with weather data, our model estimates BC at street level, achieving an $R^2$ of 0.72 and RMSE of 129.42\BCUnit. From a sustainability perspective, this work leverages resources already supported by urban infrastructure and established modeling techniques to generate information relevant to traffic emission. Obtaining BC concentration data provides actionable insights to support pollution reduction, urban planning, public health, and environmental justice at the local municipal level.
\end{abstract}


  
\maketitle

\section{Introduction}
Black carbon (BC) is produced from the incomplete combustion of fossil fuels, biomass, and biofuels. High exposures to BC adversely affect human health by increasing the risks of asthma, cardiovascular disease, and premature death in vulnerable populations (e.g., children, the elderly, and those with preexisting conditions) \cite{weichenthal2024long}. As a short-lived climate pollutant (SLCP), BC contributes to rapid warming but remains in the atmosphere for only days to decades, unlike long-lived gases such as carbon dioxide. Reducing BC, among other SLCPs, offers a way to slow global warming quickly while delivering substantial air quality and health benefits~\cite{gustafsson2016convergence}. 

Although BC emissions are primarily generated from household solid fuel use and open biomass burning, vehicle emissions are a significant contributor in urban areas \cite{weichenthal2014characterizing,hilker2019traffic}. In large cities such as Toronto, hotspots are located primarily near major roads, highways, and freight distribution centers adjacent to socially disadvantaged neighborhoods \cite{torbatian2024societal}. Despite its considerable effects on human and environmental health, data on BC pollution from traffic activities remain far less accessible to the public than data on other common pollutants such as carbon dioxide \cite{brewer2017black}.
Environmental research studies related to vehicle-based BC emissions have identified the speed and type of vehicles as key local factors, ignoring complications attributed to spatial-temporal variability on urban and regional scales \cite{weichenthal2014characterizing,dons2013street}. Specifically, urban structures and environmental factors, such as wind speed and humidity, influence particle accumulation. 

BC is typically measured using mobile air quality monitoring systems. These systems utilize Google Street View cars, with photoacoustic spectrometers~\cite{messier2018mapping} or optical microaethalometers~\cite{liu2019spatial,talaat2021mobile,torbatian2024societal} to measure pollutant concentrations. The potential for pervasive monitoring with these instruments is due to their prohibitive costs. BC concentration data from vehicles is also inaccessible because manufacturers use proprietary protocols, offering little to no publicly available data standard 
% \footnote{Vehicle emission data is generally reported for components such as carbon monoxide, nitrogen oxides, hydrocarbons, and particulate matter (PM), but precise measurements of BC are rarely available \cite{Sunwoo2025BlackCarbon}. Similarly, while there are low-cost, portable devices that provide data on $PM_{2.5}$ concentrations, they do not offer information on black carbon levels.}.
% Monitoring BC for environmental health is urgent, particularly because one of the largest urban sources stems from our (basic) transportation infrastructure. Yet, 
Conversely, intelligent transportation systems focus largely on traffic management, vehicle and driver behavior, and road safety solutions \cite{ahmad2025making}; thus, it highlights a critical gap between abundant traffic surveillance data and the scarcity of traffic emission data. 

This data gap motivates our exploration of alternative methods to estimate BC concentrations from traffic data. We hypothesize that \emph{traffic feeds can serve as a viable data source to approximate BC concentration in urban traffic conditions}. Specifically, we pose the following research questions:
\begin{itemize}
    \item[\textbf{RQ1.}] How accurately can a machine learning model trained on video-derived traffic features estimate BC concentrations?
    \item[\textbf{RQ2.}] What video-derived traffic features are most predictive of local BC concentrations?
    \item[\textbf{RQ3.}] Under what conditions and for what reasons does the model perform poorly across traffic scenarios?
\end{itemize}

Using a data collection platform that automatically integrates traffic video streams with environmental data, we conducted a field study along several common roads along Downtown Toronto to collect BC concentration readings using a microaethalometer for ground truth. We analyzed the resulting dataset with well-established computer vision (CV) models to extract characteristics of vehicle emission covariates (i.e., \texttt{vehicle type}, \texttt{vehicle count}, \texttt{stop} and accelerate behaviors on different lanes) under varying traffic conditions. Having video-derived features and API-based information from traffic and environmental data as input, we built a machine learning model to estimate BC concentrations at 30-second intervals. Our results from evaluating an XGBoost regression model yield strong predictive performance (RMSE = 129.42 \BCUnit, $R^2$=0.72), with vehicle acceleration, vehicle distance from the microaethalometer and wind speed being the most significant predictors. These findings are consistent with previous findings that BC is concentrated near the emission origin and decreases with distance. 
% It reinforces our modeling approach in estimating localized BC emissions from vehicles in traffic, which diminish with increasing distance from the source. 

Our implementation highlights the technical feasibility of leveraging traffic video recording and established CV techniques to accurately determine a set of vehicle emission covariates as significant predictors of approximating BC concentrations in traffic situations. However, our work clearly requires longitudinal data collection to capture seasonal variations. It should be emphasized that the focus of this study is on the detection framework itself, which encompasses the development of our data collection platform and techniques for feature extraction and modeling. This approach is not intended to replace high-precision microaethalometers, as they remain necessary to provide ground-truth measurements. Nonetheless, by continuing to use traffic data in novel ways, we can move beyond traffic monitoring insights to inform end-users of health-related information. From a sustainability perspective, this work rests on two key ideas:

\begin{enumerate}
    \item It builds on robust field-tested machine learning methods that form the backbone of deployed traffic safety systems and common urban technologies. %The contribution is not a new technical implementation, but rather the ability to extract the fullest possible value from data that are already circulating in our urban environment.
    \item Using widely available infrastructure and abundant traffic feed data, it enables cities to address data gaps that cannot afford dedicated equipment. 
    % \item It generates estimates on BC concentration in traffic to reveal where pollution reductions are most needed. Given the abundance of traffic feed data, our work allows information to be used in both forward-looking projections and retrospective analyses, making the data more versatile for guiding city planning, policy interventions and public health measures.
\end{enumerate}

While some location-specific calibration using limited ground-truth measurements may be necessary, this requirement is far less costly than deploying a full sensor network, making the approach practical and scalable. Thus, our research directly supports the United Nations Sustainable Development Goals \cite{assembly2015transforming}. 
% Unlike the more regionally uniform distribution of $PM_{2.5}$, BC exhibits strong spatial heterogeneity, with concentrations rising sharply near major roadways and heavy-duty traffic corridors \cite{zhang2019black,de2019relationship,minet2018development}. 
The fine-scale spatial and temporal variability of BC emissions makes it a particularly relevant metric for assessing intra-urban disparities in exposure \cite{savadkoohi2023variability,saeedi2025urban,jaafar2025refining}. At the same time, marginalized and low-income communities are often located near high-traffic environments and industrial corridors, resulting in disproportionate exposure burdens. 
Our work enables both forward-looking projections and retrospective analyses, making the data more versatile to guide city planning, policy interventions, and public health measures~\cite{batisse2025examining,northcross2020monitoring}. In doing so, our efforts aim to reduce the disproportionate burden on disadvantaged communities (SDG 10), inform sustainable urban development (SDG 11) and population health (SDG 3), and accelerate climate action by targeting SLCPs (SDG 13). 

We extend the impact of our work by providing our data collection system, dataset, and models as publicly available resources via \url{https://github.com/sensAILabs/BCTraffic}. Descriptions of these resources are also in the Appendix.
\section{Related Work}
We review work on vehicle-related emissions, the environmental factors that influence them, and estimation methods for BC concentration, including ML approaches.

\subsection{Data Collection for Black Carbon} 
Portable instruments on mobile platforms are often utilized to characterize this spatiotemporal variability and the actual levels of BC \cite{talaat2021mobile}. Measurement of BC concentration requires optical analyzers such as a microaethalometer, which measures the attenuation of an 880-nanometer radiation beam transmitted through a filter strip \cite{liu2019spatial}. The device costs approximately USD\$6,000 and requires expert knowledge. Such device inaccessibility has led to significant amounts of missing data in BC reporting and hinders public health education and the development of evidence-based environmental policies. For example, the Canada Air Pollutant Emissions Inventory (APEI) approximates the BC concentration resulting from Transportation and Mobile Equipment through air transportation, domestic marine navigation, fishing and military, and rail transportation reporting \cite{APEI}. These reports do not include elements with which the general public is likely to encounter or engage, such as those from on-road vehicles in urban and suburban traffic conditions. In addition, individual vehicle manufacturers have a bespoke testing protocol; specifically, manufacturers report the total particulate mass from which BC must be inferred indirectly \cite{Sunwoo2025BlackCarbon}.

\subsection{Emissions from Local Vehicle Sources}
Research has reported increasing concerns about the negative impacts of transportation on BC pollution. Specifically, near-road air quality in urban areas can fluctuate significantly over short distances and time frames \cite{parvez2019hybrid}. BC concentration is strongly influenced by vehicle type, speed, and traffic density. A study by Zheng \emph{et al.} measured BC concentration from eight light-duty passenger vehicles (LDPVs) with different engine technologies using a dynamometer and a microaethalometer (the same BC device we utilized for our study). The results showed that the BC emission factors varied significantly depending on the vehicle type and its engine. Diesel engines emitted between 3.6 to 91.5 mg/km, gasoline direct injection (GDI) engines produced 7.6 mg/km, and gasoline port-fuel injection (PFI) engines had the lowest emission rates, ranging from 0.13 to 0.58 mg/km \cite{zheng2017characteristics}. Gasoline PFI vehicles exhibited peak BC emissions during cold starts and aggressive high-speed driving, which contributed to elevated ambient BC concentrations. On the other hand, heavy-duty vehicles (HDVs) such as buses and trucks contributed disproportionately to both emissions and concentration levels \cite{zhang2019black} The influence of vehicle speed should be considered along with vehicle type \cite{banweiss2009measurement}. At very low speeds (~30 km/h), common in urban congestion, BC emissions tend to be high due to frequent stop-and-go driving, leading to incomplete fuel combustion due to frequent acceleration and deceleration \cite{dons2013street,zhang2019black}. At higher speeds (50–70 km/h), emissions tend to be lower, as combustion is more efficient under steady driving conditions. Large cities like Toronto and Montreal are predominantly characterized by clusters of BC sources concentrated around transportation infrastructure and dense urban cores \cite{batisse2025examining}. Suburban areas, on the other hand, while generally experiencing lower overall traffic volumes, can exhibit higher per capita emissions due to a greater reliance on personal vehicles and a more limited access to public transportation options.

\subsection{Traffic Air Pollution Mobile Monitoring}
Mobile setups for outdoor air quality monitoring systems typically consist of Google Street View cars as the mobile sampling platform, and are equipped with 1-Hz instruments (e.g., gas analyzers) to measure the concentrations of pollutants \cite{messier2018mapping}. Several studies aimed to understand BC pollution in different cities, such as Cairo, Shanghai and Toronto, employed microaethalometer (AE51, AethLabs) for BC data collection \cite{liu2019spatial,talaat2021mobile,torbatian2024societal}. Generally, BC measurements are challenging because they are influenced by spatial-temporal variability on the \emph{local}, \emph{urban}, and \emph{regional} scales \cite{weichenthal2014characterizing}. In the example of near-road air quality, BC amounts are determined by attributes of local sources (e.g., emission from a truck differs from a bus) and impacts of wind speed and humidity from local and nearby regions. 

\begin{figure*}[h!]
    \centering
    \includegraphics[width=.75\linewidth]{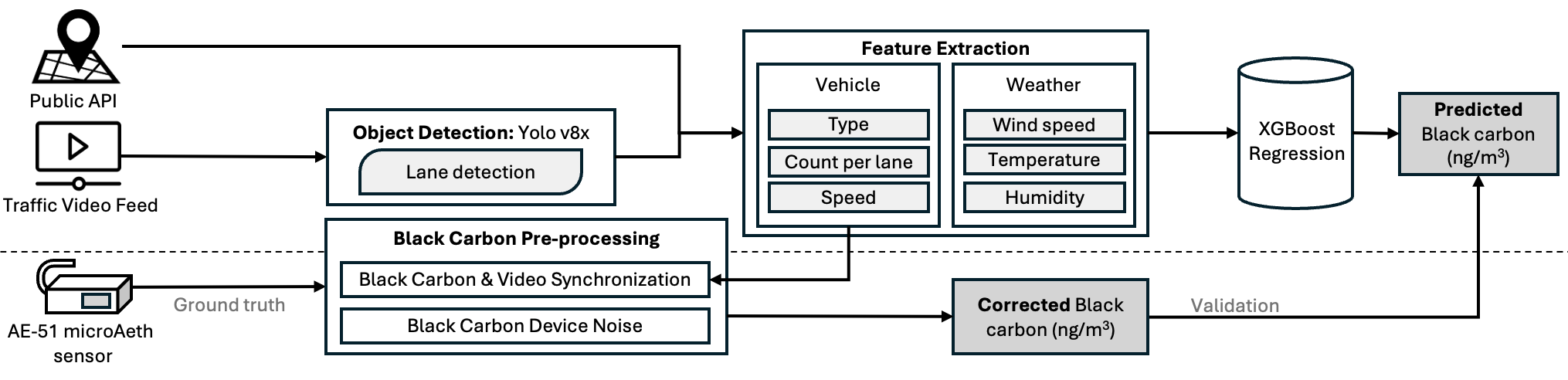}
    \caption{End-to-end system overview for BC estimation with computer vision and machine learning techniques.}
    \label{fig:systemOverview}
\end{figure*}

Using Land Use Regression, this method uses environmental data to identify how different land uses influence BC concentration. Recent studies reported that ML models, such as Random Forests \cite{krecl2019modelling} and Kernel-based Regularized Least Squares \cite{weichenthal2016land}, outperform traditional regression models in capturing complex nonlinear relationships related to the built environment. Separately, studies on vehicle identification have proposed CV techniques, implemented with deep learning (DL) models to predict vehicle speed using traffic surveillance \cite{hua2018vehicle}, vehicle type using sound \cite{dawton2020initial}, and improved detections under varying weather conditions using LiDAR \cite{qian2021robust}. However, such analysis is based on static features of the built environment. Our goal is to incorporate real-time vehicle behaviors and conditions that directly drive emissions.

\subsection{Why Traffic Video Feeds?} Air quality analysis typically relies on a combination of regulatory-grade monitoring stations, which provide high accuracy. A single regulatory-grade station can cost in the range of several hundred thousand dollars to install and maintain, making pervasive coverage across cities or rural areas impractical. Low-cost fixed sensors and mobile sensing platforms can complement fixed setups. However, a study by deSouza \emph{et al.} on the distribution of low-cost air quality sensors within the U.S. reported these sensors were more available in higher-income, predominantly White areas \cite{desouza2021distribution}. At the same time, CNBC estimated over 770 million surveillance cameras installed worldwide, with 18\% located in the Americas and 54\% concentrated in China \cite{Samsel2019OneBillionSurveillance}. Today, nearly every major city has invested in intelligent traffic management or law enforcement surveillance, with cameras and sensors continuously generating vast data streams. These systems undeniably provide powerful capabilities for monitoring and regulating traffic violations and driver behavior. However, the disproportionately higher deployments of surveillance cameras in low-income neighborhoods under the banner of traffic and community safety have raised important equity concerns \cite{SuttonTilahun2022ChicagoCameraTicket}. The abundance of traffic cameras, while enhancing safety and enforcement, not only raises equity concerns but also highlights a critical gap. Data on traffic-related emissions remain scarce \cite{ahmad2025making}. This disconnect prompts a critical question: \emph{\textbf{How can traffic data already available in the environment draw inferences about black carbon concentration from traffic?}}\\

\noindent \textbf{\emph{Key Takeaway:}} Collectively, these reports support our direction to explore traffic video data as a potential source, as many environmental studies rely on a fixed set of vehicle emission covariates we believe can be captured using CV techniques. First, \emph{stop-and-go} conditions are most evident during periods of heavy traffic congestion. Second, \emph{heavy-duty vehicles} have a disproportionate impact and short spikes are strongly influenced by vehicle states such as idling and acceleration. Environmental features also matter at short time scales: \emph{wind speed} influences local concentration. Although CV and ML research has advanced many areas of traffic monitoring and surveillance, prior work has not examined how the infrastructure and techniques can be used to predict BC from traffic.
\section{Proposed System}
\autoref{fig:systemOverview} illustrates the overview of our proposed system. We estimate key vehicular emission covariates by applying CV techniques to analyze traffic video recordings and extract local factors (i.e., vehicle and traffic-related characteristics).
These features are used to approximate BC concentration as measured by a microaethalometer. 

\subsection{Video-Derived Vehicle Features}
We extract vehicle-related information from video data using a pre-trained YOLOv8 model \cite{varghese2024yolov8}, which is able to accurately determine vehicle characteristics (e.g., cars from trucks). However, extracting these vehicle emission covariates involves two non-trivial challenges. The first relates to distinguishing whether vehicles are closer or farther away from the deployed microaethalometer. The other requires determining a stop or idle as such actions emit high BC concentration when vehicles accelerate.

\subsubsection{Assumptions}
Several aspects of our implementation are based on underlying assumptions. First, the vehicle lanes in our dataset are relatively straight (see \autoref{fig:streetLayout} for street illustrations). Although established CV techniques can accommodate curved or non-linear lanes, ours does not adopt these complex approaches. Second, the camera placement limits our field of view, occasionally resulting in partial occlusion of full traffic flow. Standard traffic-surveillance setups do not face this constraint, and we mitigate this limitation by pulling traffic density and speed data from the TomTom API \cite{tomtom_traffic_api}.

\subsubsection{Detecting Vehicles by Lane}
\begin{figure}
    \centering
	\includegraphics[width=\linewidth]{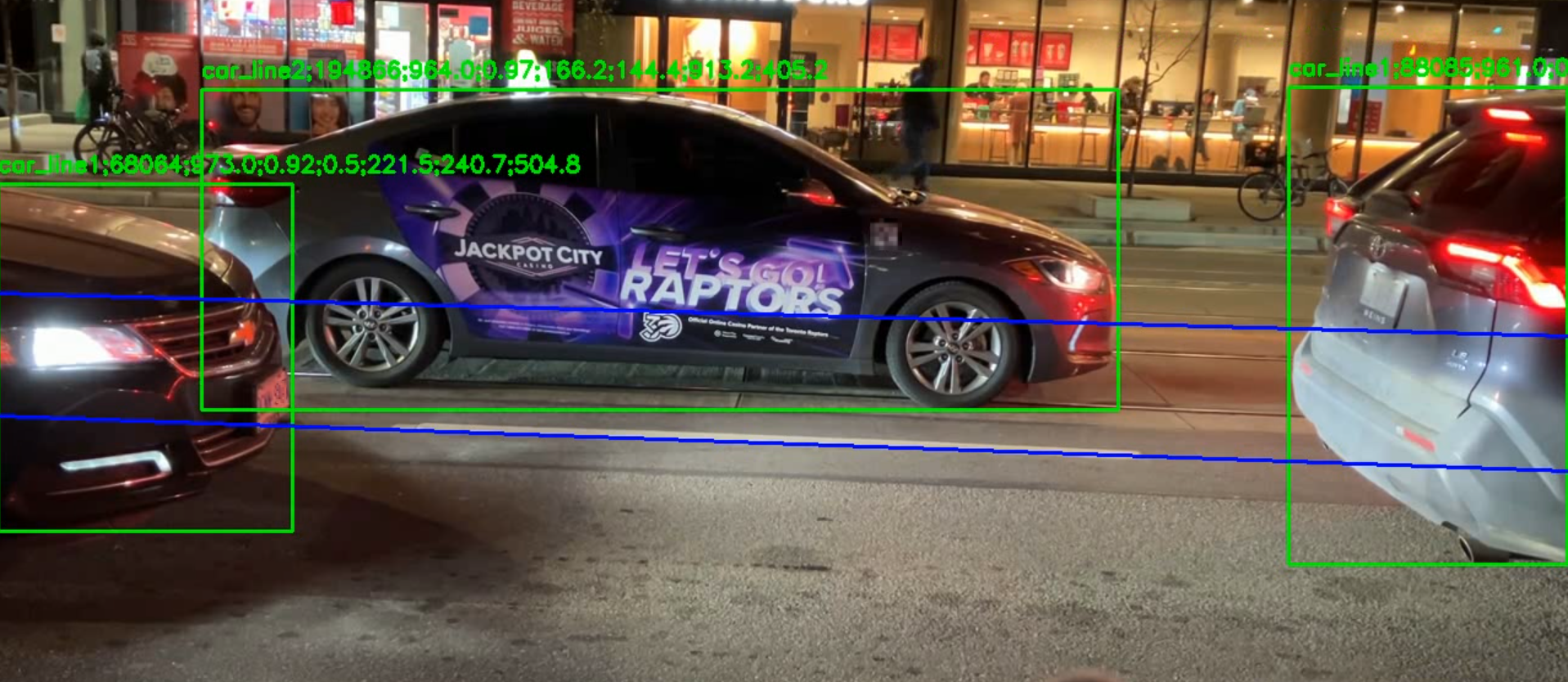}
    \captionof{figure}{Result of applying YOLO and Hough Transform to determine the type of vehicle and distance from the microaethalometer.}
    \label{fig:carCV}
\end{figure}

Our approach requires differentiating vehicles by traffic lanes, where vehicles along the first lane closest to our setup are expected to have the most direct impact on BC measurements. However, this capability is not inherently supported by YOLO. Hence, we use the Hough Transform, a well-established CV technique to detect straight lines \cite{duda1972use}. As per \autoref{fig:carCV}, it employs Canny Edge Detection to extract edges, producing lane lines frame-by-frame. Once the lanes are identified, each detected vehicle is assigned a lane by comparing the vehicle's centroid in the boundary box with the lane boundary lines. This segmentation allows us to count vehicles in each lane separately.

\subsubsection{Vehicle Stop-and-Idle and Count Estimation} To differentiate stopped or idle vehicles from those that only decelerate, we used a heuristically driven approach. First, we extended the vehicle detection module with a threshold-based method that analyzes motion across consecutive frames. By tracking positional shifts and estimating speed, we classify a vehicle as “stopped” if its speed remains near zero for at least four seconds. In addition, we incorporated traffic density and speed data from the TomTom API to help estimate vehicle counts and overall flow conditions, compensating for the limited field of view from our camera placement. Together, these heuristics allowed us to better distinguish stationary vehicles from those that simply moved slowly.

%To differentiate stopped or idle vehicles from those that merely decelerate without stopping, we employed a heuristically driven approach. Specifically, we extended the vehicle detection module with a threshold-based method to analyze motion across consecutive frames. By tracking positional shifts and computing speed changes, we classify a vehicle as ``stopped'' if its speed remains near zero for at least four seconds. This approach enables us to distinguish stationary vehicles, responsible for disproportionately high BC concentration, from those that move slowly but continuously.

\subsection{Black Carbon Pre-processing Module}
Microaethelometers sample BC concentration at a fixed resolution of 10–30 seconds \cite{minet2018development}. 
While we rely on this data as the ground truth for our models, we pre-process it to handle two key challenges: the presence of noise caused by faulty device calibration, and unsupported synchronization between the BC ground truth and extracted video features.

\subsubsection{Challenge \#1: Microaethalometer Data Noise}
\label{sec:noiseStrategy}
To preserve data integrity while minimizing the effect of noise, we apply the Optimized Noise-Reduction Algorithm (ONA), a filtering technique by Hagler \emph{et al.} that applies adaptive time-averaging to BC readings, adjusting the averaging window based on the incremental light attenuation detected by the instrument’s internal filter \cite{hagler2011post}. Separately, we remove outliers, data points that fall outside the 95\% confidence interval. Our method distinguishes between global and local trimming; that is, local trimming preserves the unique distribution in each dataset, while global trimming applies the same criteria across all datasets. %Applying the confidence interval (CI) trimming essentially removes the top and bottom 2\% of our training dataset.

\subsubsection{Challenge \#2: Synchronization between Black Carbon and Video Data}
To address the temporal misalignment between BC ground truth data and current traffic density, we identified the optimal time shift that best aligns the two readings. Specifically, we transformed both signals into the frequency domain using the Discrete Fourier Transform (DFT), where time shifts are reflected as phase differences. Then, we computed the cosine similarity across varying phase offsets to determine the time shift that yielded the highest alignment. Specifically, let \( x[n] \) and \( y[n] \) represent the two signals. The DFT is applied to both signals, where \( X[k] \) and \( Y[k] \) are the representations of the frequency domain, \( j \) facilitates the representation of both magnitude and phase information of the signal components, and \( k \) indexes the discrete frequency components, as follows:
\[
X[k] = \sum_{n=0}^{N-1} x[n] e^{-j 2 \pi \frac{kn}{N}}, \quad Y[k] = \sum_{n=0}^{N-1} y[n] e^{-j 2 \pi \frac{kn}{N}}
\]

To identify the optimal time shift \( \Delta t_{\text{optimal}} \), we compute the cosine similarity between the frequency domain representations for various shifts, where \( \langle \mathbf{X}, \mathbf{Y} \rangle \) is the dot product of the cosine and sine waves of the DFT vectors and \( \|\mathbf{X}\| \) and \( \|\mathbf{Y}\| \) are their Euclidean norms. 
\[
\text{Cosine Similarity} = \frac{\langle \mathbf{X}, \mathbf{Y} \rangle}{\|\mathbf{X}\| \|\mathbf{Y}\|}
\]

With the components separated as $X_{\text{cos}} = \text{Cos}(X[k])$, $X_{\text{sin}} = \text{Sin}(X[k])$, $Y_{\text{cos}} = \text{Cos}(Y[k])$, $Y_{\text{sin}} = \text{Sin}(Y[k])$, respectively, we compute the cosine similarity between these components as follows:
\[
\text{Cosine Similarity} =
\frac{\langle X_{\cos}, Y_{\cos} \rangle + \langle X_{\sin}, Y_{\sin} \rangle}
{\sqrt{\|X_{\cos}\|^2 + \|X_{\sin}\|^2} \;\;
 \sqrt{\|Y_{\cos}\|^2 + \|Y_{\sin}\|^2}}
\]

Where \( \mathcal{S} \) is the set of candidate shifts within a reasonable range, we experimentally test various time shifts to identify the most optimal shift, \( \Delta t_{\text{optimal}} \), which maximizes the cosine similarity between the representations of the frequency domain, thus aligning the time domain signals.

\[
\Delta t_{\text{optimal}} = \arg\max_{\Delta t \in \mathcal{S}} \left( \text{Cosine Similarity}(X[k], \mathcal{F}\{y[n - \Delta t]\}) \right)
\]

\begin{figure}
    \centering
    \includegraphics[width=.8\linewidth]{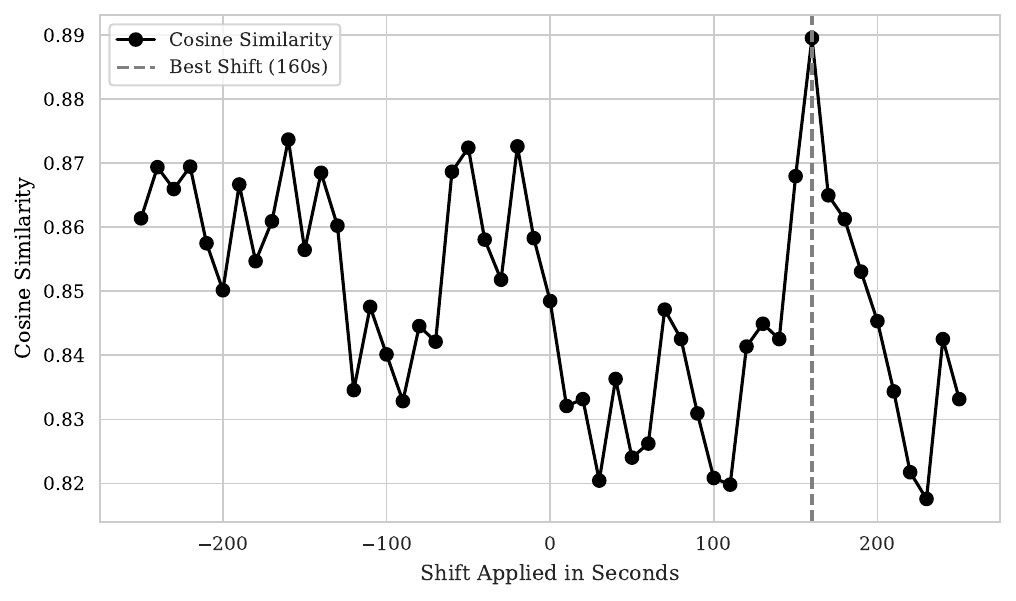}
    \captionof{figure}{Cosine similarity between Fourier transformed BC vector and Fourier transformed \emph{TotalVehicle} vector. The corrected BC signal is determined by the shift that maximizes cosine similarity (160 seconds).}
    \label{fig:maxCosineSimilarity}
\end{figure}

\begin{figure}
    \centering
    \includegraphics[width=.8\linewidth]{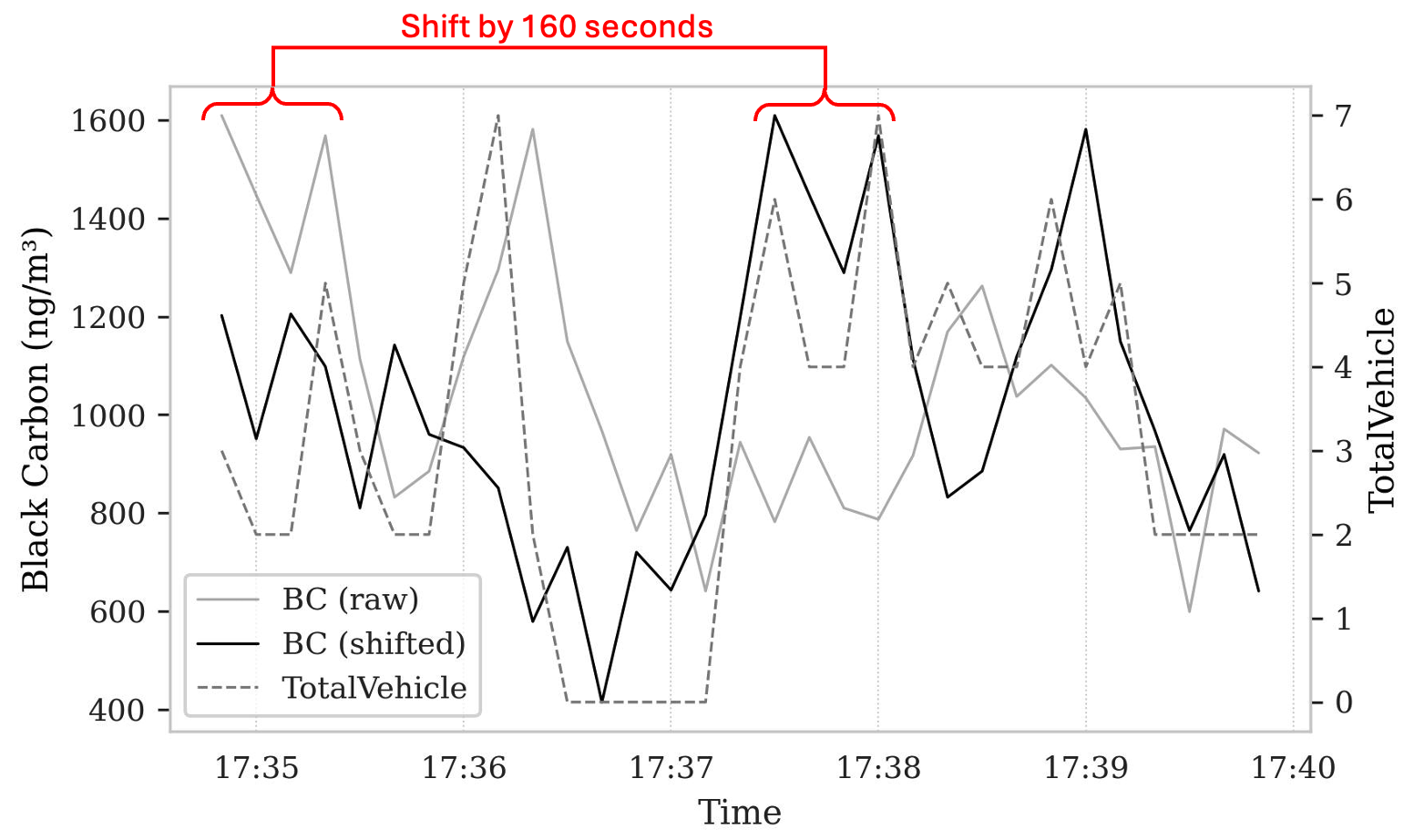}
    \captionof{figure}{Original BC signal (gray) plotted alongside vehicle counts (dotted line) reveals a temporal misalignment. The corrected BC signal (black), shifted by 160 seconds, shows alignment with vehicle activity. }
    \label{fig:temporalMisalignment}
\end{figure}

By applying this shift as shown in \autoref{fig:maxCosineSimilarity}, we achieved corrected BC-FFT signals that improve the alignment between BC data and vehicle activity, as shown in \autoref{fig:temporalMisalignment}.

\subsection{Black Carbon Regressor Model}
\autoref{tab:summaryfeatures} presents the complete set of features extracted from our pipeline, consisting of vehicle emission covariates and environmental features. Our analysis, detailed in Section \ref{sec:caseAnalysis}, further refines this set for model prediction. We also obtained environmental features, such as wind speed, temperature, and humidity, from publicly available APIs \cite{Zippenfenig_Open-Meteo}.

\begin{table}[h!]
    \centering
    \caption{List of vehicular-based and environmental features for black carbon prediction.}
    \scalebox{.85}{
    \begin{tabular}{l|l}\hline
    \textbf{\makecell[l]{Feature}} & \textbf{Description} \\\hline
    \makecell[l]{$TotalVehicle_{t}$} & \makecell[l]{Count of all vehicles at time $t$} \\\hline
    \makecell[l]{$LDPV_{l}$} & \makecell[l]{Count of LDPV in lane $l$} \\\hline
    \makecell[l]{$HDV_{l}$} & \makecell[l]{Count of HDV in lane $l$} \\\hline

    \makecell[l]{$StopLDPV_{l}$} & \makecell[l]{Count of LDPVs stopped in lane $l$} \\\hline
    \makecell[l]{$StopHDV_{l}$} & \makecell[l]{Count of HDVs stopped in lane $l$}\\\hline
    
    \makecell[l]{\emph{his\_humid}} & \makecell[l]{Historical humidity (2 min prior, \%)}  \\\hline
    \makecell[l]{\emph{his\_temp}} & \makecell[l]{Historical temperature (2 min prior, °C)} \\\hline
    \makecell[l]{\emph{his\_wind}} & \makecell[l]{Historical wind speed \\(10 meter above the earth, km/s)} \\\hline

    \end{tabular}}
    % \\\vspace{0.05in}\footnotesize{First lane: <3~m away; Second lane: 3–6~m away; Third lane: >6~m away}
    \label{tab:summaryfeatures}
\end{table} 

\begin{figure}
    \centering
    \includegraphics[width=.8\linewidth]{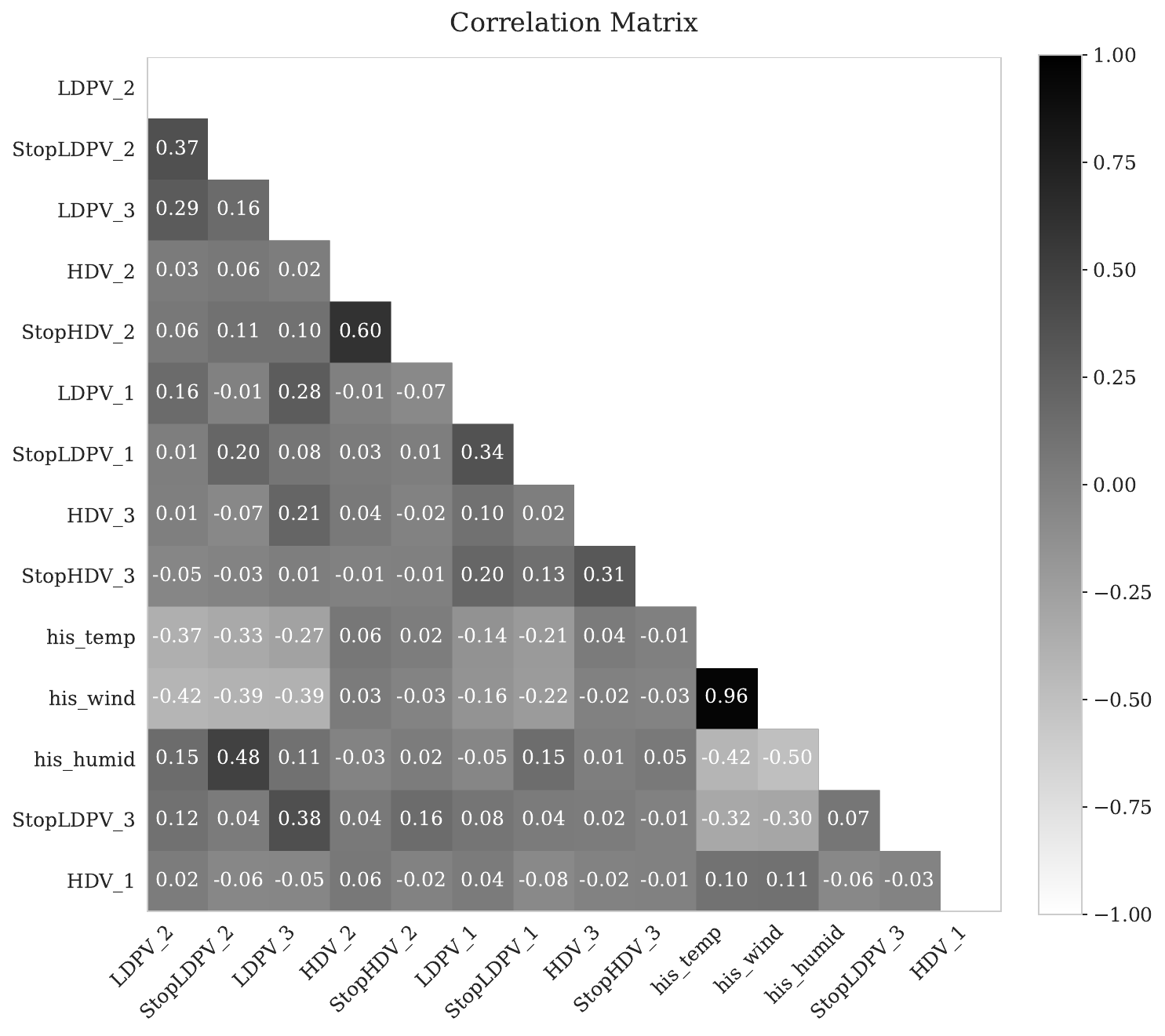}
    \captionof{figure}{Correlation matrix. We excluded highly correlated features above 0.70.}
    \label{fig:corrFeatures}
\end{figure}
Our primary objective is to determine whether video-based generated characteristics can effectively predict BC concentrations, aligning with environmental air quality research findings that demonstrate the significance of vehicle emission covariates in BC pollution. We formulate this as a supervised learning problem because the BC measurements serve as ground truth labels to train a model that can directly map input features and BC concentration. 

Our model uses the XGBoost algorithm \cite{chen2016xgboost} using 75\% of the dataset, as summarized in \autoref{tab:dataset}. XGBoost is well known for its computational efficiency and model performance \cite{ester2022xgboost}. Its regularized gradient-boosting framework builds trees sequentially to minimize training loss while controlling overfitting through shrinkage; after tuning, the best-performing hyperparameters were a learning rate ($learning\_rate$) of 0.05, a maximum tree depth ($max\_depth$) of 5, and 50 trees ($n\_estimators$). It is trained on vehicle and environment features, and highly correlated features are excluded (see \autoref{fig:corrFeatures}).
\section{Study}
\begin{figure}[tb!]
    \centering
	\includegraphics[width=.8\linewidth]{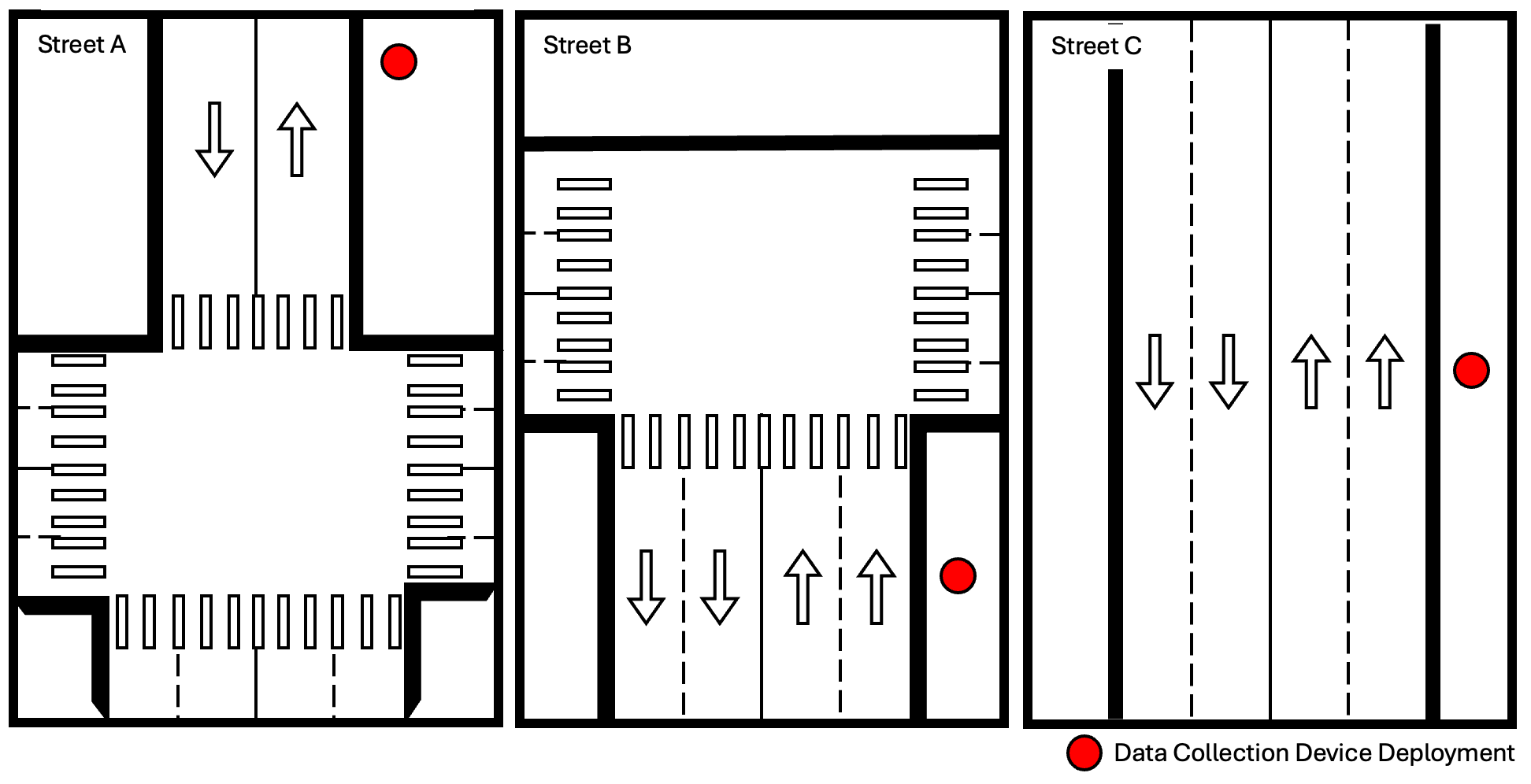}
    \captionof{figure}{Three streets layouts for data collection.}
    \label{fig:streetLayout}
\end{figure}

This study is approved by our university’s research ethics board. We collected traffic video and BC concentration data from three different streets in Downtown Toronto, a densely populated urban area with some of the worst traffic congestion globally \cite{batisse2025examining}. \autoref{fig:streetLayout} illustrates three distinct street layouts from which the data were collected.

\subsection{Data Collection System Setup}
\begin{figure}[tb!]
    \centering
	\includegraphics[width=.6\linewidth]{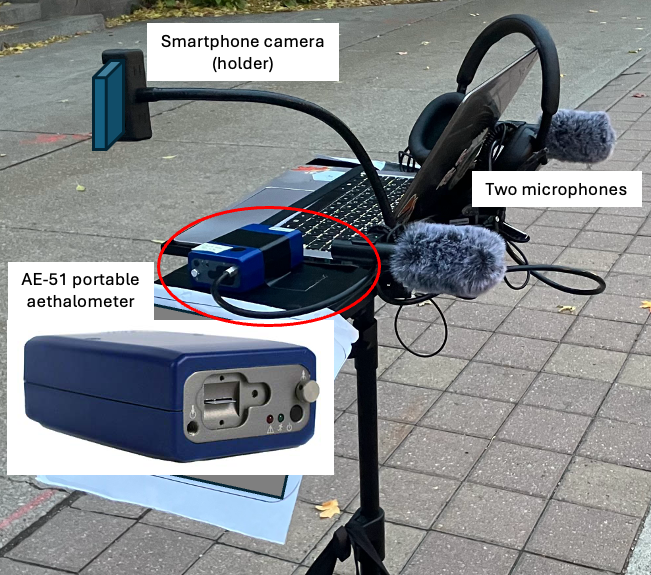}
    \captionof{figure}{Device setup for data collection.}
    \label{fig:station}
\end{figure}

As illustrated in \autoref{fig:station}, we designed a setup consisting of a microaethalometer sensor to measure BC concentration, a microphone to capture audio signals and a smartphone camera to record video footage. This setup mimics the one used by Dawton \emph{et al.} for roadside vehicle identification~\cite{dawton2020initial}. Separately, we developed an online data collection application (see Appendix \autoref{fig:dashboard}) using a combination of FastAPI for the server side and Jinja2 for templating on the client side. We integrated open-source APIs, including OpenMeteo and OpenStreetMap \cite{Zippenfenig_Open-Meteo, OpenStreetMap}. For real-time traffic updates, we utilized the TomTom Traffic API \cite{tomtom_traffic_api}. This setup allowed our research team to configure custom data collection studies combining time periods, environmental conditions, and traffic variables for analysis. 

\begin{table*}[t!]
    \centering
    \caption{Summary of data collected.}
    \scalebox{0.85}{
    \begin{tabular}{l|c|c|c|c|c|c|c} \hline
    \textbf{} & \textbf{Location} & \textbf{Duration}  & \textbf{\makecell[c]{LDPV}} & \textbf{\makecell[c]{HDV}} & \textbf{Wind} & \textbf{Humidity} & \textbf{BC (in \BCUnit)} \\ \hline

    \textbf{Set$_{A}$} & \makecell[c]{Lat: 43.658 \\ Lon: -79.395}  
 & 49.0 mins  
        & \makecell[c]{97.4\%, $n$=736} 
        & \makecell[c]{2.6\%, $n$=20}
        & \makecell[c]{$\mu$=15.59, $\sigma$=0.10, \\$\min$=15.5, $\max$=15.7} 
        & \makecell[c]{$\mu$=68.86, $\sigma$=0.99, \\$\min$=68.0, $\max$=70.0} 
        & \makecell[c]{$\mu$=828.03, $\sigma$=219.51, \\$\min$=427, $\max$=1508} \\ \hline

    \textbf{Set$_{B}$} & \makecell[c]{Lat: 43.660 \\ Lon: -79.397} & 32.5 mins  
        & \makecell[c]{95.1\%, $n$=442} 
        & \makecell[c]{4.9\%, $n$=23} 
        & \makecell[c]{$\mu$=27.03, $\sigma$=1.57, \\$\min$=24.5, $\max$=28.0} 
        & \makecell[c]{$\mu$=66.28, $\sigma$=0.45, \\$\min$=66.0, $\max$=67.0} 
        & \makecell[c]{$\mu$=729.82, $\sigma$=260, \\$\min$=-242, $\max$=1623} \\ \hline

    \textbf{Set$_{C}$} & \makecell[c]{Lat: 43.659 \\ Lon: -79.393} & 32.5 mins 
        & \makecell[c]{99.5\%, $n$=551} 
        & \makecell[c]{0.5\%, $n$=3} 
        & \makecell[c]{$\mu$=5.70, $\sigma$=0.69, \\$\min$=4.9, $\max$=6.3} 
        & \makecell[c]{$\mu$=67.29, $\sigma$=1.49, \\$\min$=66.0, $\max$=69.0} 
        & \makecell[c]{$\mu$=391, $\sigma$=191, \\$\min$=93, $\max$=973} \\ \hline

    \textbf{Set$_{D}$} & \makecell[c]{Lat: 43.659 \\ Lon: -79.394} & 10.0 mins
        & \makecell[c]{100\%, $n$=90} 
        & \makecell[c]{0\%, $n$=0} 
        & \makecell[c]{$\mu$=9.20, $\sigma$=0.00, \\$\min$=9.2, $\max$=9.2} 
        & \makecell[c]{$\mu$=78.00, $\sigma$=0.00, \\$\min$=78.0, $\max$=78.0} 
        & \makecell[c]{$\mu$=502, $\sigma$=269, \\$\min$=167, $\max$=1391} \\ \hline

    \end{tabular}}
    \label{tab:dataset}
\end{table*}

% In addition, it maintains a repository of previous experiments, allowing the team to review, update, or expand on past analyses. 

\subsection{Dataset}
\label{sec:dataset}
\autoref{tab:dataset} summarizes the details of our dataset. It consists of short-term BC concentration measurements collected from four road junctions across different days, within a limited geographical area of our university. Thus, our data resulted in a highly constrained and heterogeneous sample size. With a total of 124 minutes, it is evidently too small to support a robust spatio-temporal evaluation. The constraint was dictated by logistical issues from loaning a microaethalometer within a restricted period, preventing an extensive and balanced data collection campaign. 

\begin{figure}
    \centering
	\includegraphics[width=.85\linewidth]{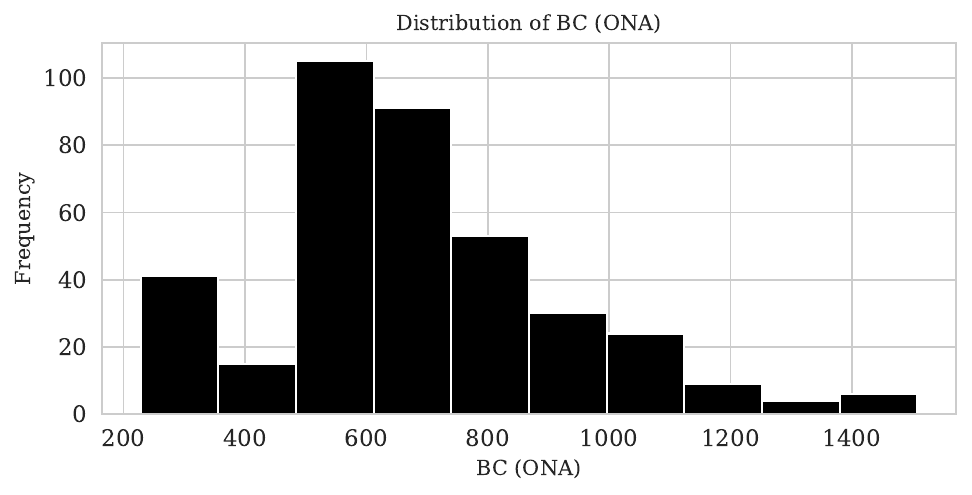}
    \captionof{figure}{Distribution of BC ground truth readings from the microaethalometer after applying the Optimized Noise-Reduction Algorithm (ONA). }
    \label{fig:BCdistribution}
\end{figure}

The composition of vehicles was predominantly light duty in all sets, accounting for approximately 98\% of the time, with HDVs constituting a minor fraction. Weather conditions varied throughout the study period, ranging from cloudy skies with strong winds to milder breezes and occasional snowfall. Wind speeds fluctuated between low and moderate, while humidity levels were predominantly in the moderate to high range, with periods of very high humidity. BC concentrations measured by the microaethalometer peaked at 1623\BCUnit, with the lowest values attributed to device miscalibration. As per \autoref{fig:BCdistribution}, the average BC concentration was 685\BCUnit, with a standard deviation of ±249 after noise reduction.

\section{System Evaluation}
\label{sec:evaluation}
\subsubsection{Objective} Our modeling objective is to evaluate how well vehicle-related features explain BC concentrations, rather than to demonstrate strong spatio-temporal generalization. Data from all four collection periods were combined before stratified sampling of 75\% train and 25\% test to achieve two aims. First, it ensures the model is trained on the full range of BC concentrations and traffic-related feature variations observed across all sites. Second, it allows the test set to evaluate the model’s ability to interpolate pollutant levels within the conditions already represented in the data. As a supplementary evaluation, we consider a windowed stratified split that respects the temporal ordering within each dataset by grouping samples into a fixed-length time window of 15 minutes. This ensures that all samples within a window are kept together and that training data precedes test data within each dataset. The added approach helps shed light on temporally subsequent observations. In our experiments, we removed outliers from the training data, but the test set remained unchanged. 

\subsubsection{Performance Metric} Models are evaluated using $R^{2}$, Mean Absolute Error (MAE), and Root Mean Squared Error (RMSE). While $R^{2}$ measures variability capture and MAE provides average error magnitude, we prioritize RMSE as the primary metric because it provides a more interpretable measure of predictive error, penalizes large errors and remains in the target unit of \BCUnit.

\begin{table} [h!]
    \centering
    \caption{Model evaluation using different algorithms. XGBoost regression yields the best performance.}
    \scalebox{0.8}{
    \begin{tabular}{l|c|c|c} \hline
     & \makecell[c]{\textbf{Train}\\RMSE / MAE / $R^{2}$} & %\textbf{Train MSE} & 
     \makecell[c]{\textbf{Test}\\RMSE / MAE / $R^{2}$} & %\textbf{Test MSE} & 
     \textbf{p} \\ \hline

    LR (baseline)   & 184.70 / 146.83 / 0.34   & %34115.48  & 
    211.66 / 163.56 / 0.25 & %44799.06    & 
    -   \\ \hline

    SVR      & 216.20 / 155.23 / 0.10   & %46743.69  & 
    238.49 / 170.78 / 0.05   & %56876.81    & 
    $p>.01$ \\ \hline

    RF        & 82.71 / 62.01 / 0.87    & %6840.17   & 
    131.86 / 91.70 / 0.71   & %17387.02    & 
    $p<.001$         \\ \hline

    GB              & 77.31 / 58.15 / 0.88    & %6990.81   & 
    136.76 / 93.53 / 0.68   & %19447.89    & 
    $p<.001$     \\ \hline
    
    \rowcolor{LightGrey}
    XGBoost    & 95.68 / 71.57 / 0.82   & %9155.43   & 
    129.42 / 91.11 / 0.72   & %16748.60    & 
    $p<.001$    \\ \hline

    \rowcolor{LightGrey}
    \makecell[l]{XGBoost\\(windowed) }& 82.25 / 51.88 / 0.91& %9155.43   & 
    119.93 / 93.25 / 0.30  & %16748.60    & 
    $p<.001$    \\ \hline
    \end{tabular}}
    \label{tab:resultsAlgo}
\end{table}

\begin{table} [h!]
    \centering
    \caption{XGBoost regression yields best performance with Optimized Noise-Reduction Algorithm (ONA) and removing outliers locally in each dataset.}
    \scalebox{0.8}{
    \begin{tabular}{l|c|c|c} \hline
    & \makecell[c]{\textbf{Train}\\RMSE / MAE / $R^{2}$} & %\textbf{Train MSE} & 
     \makecell[c]{\textbf{Test}\\RMSE / MAE / $R^{2}$} & %\textbf{Test MSE} & 
     \textbf{p} \\ \hline

    \makecell[l]{None\\(baseline)} & 158.93 / 123.52 / 0.67   & %25259.00  & 
    212.47 / 158.42 / 0.49& %45143.92  & 
    -\\ \hline

    ONA & 121.83 / 86.40 / 0.76   & %14841.89  & 
    140.10 / 99.43 / 0.67  & %19627.79  & 
    p > .05 \\ \hline

    Trim (global) & 137.27  / 111.46 / 0.65  & %18842.19  & 
    210.78  / 153.43 / 0.50  & %44428.09  & 
    p > .1 \\ \hline

    Trim (local) & 111.02 / 84.04 / 0.79   & %12324.98  & 
    207.22 / 144.58 / 0.52 & %42939.41  & 
    p > .05  \\ \hline

    \makecell[l]{ONA+Trim\\(global)} &  75.25 / 56.72 / 0.88 & %5662.92  & 
    135.57 / 94.20 / 0.69 & %18378.99  & 
    p < .05 \\ \hline

    \rowcolor{LightGrey}
    \makecell[l]{ONA+Trim\\(local)}  &  95.68 / 71.57 / 0.82  &  %9155.43  & 
    129.42 / 91.11 / 0.72 & %16748.60  & 
    p < .05 \\ \hline

    \end{tabular}}
    \label{tab:noiseResult}
\end{table}

\begin{figure*} [h!]
    \centering
    \includegraphics[width=.8\linewidth]{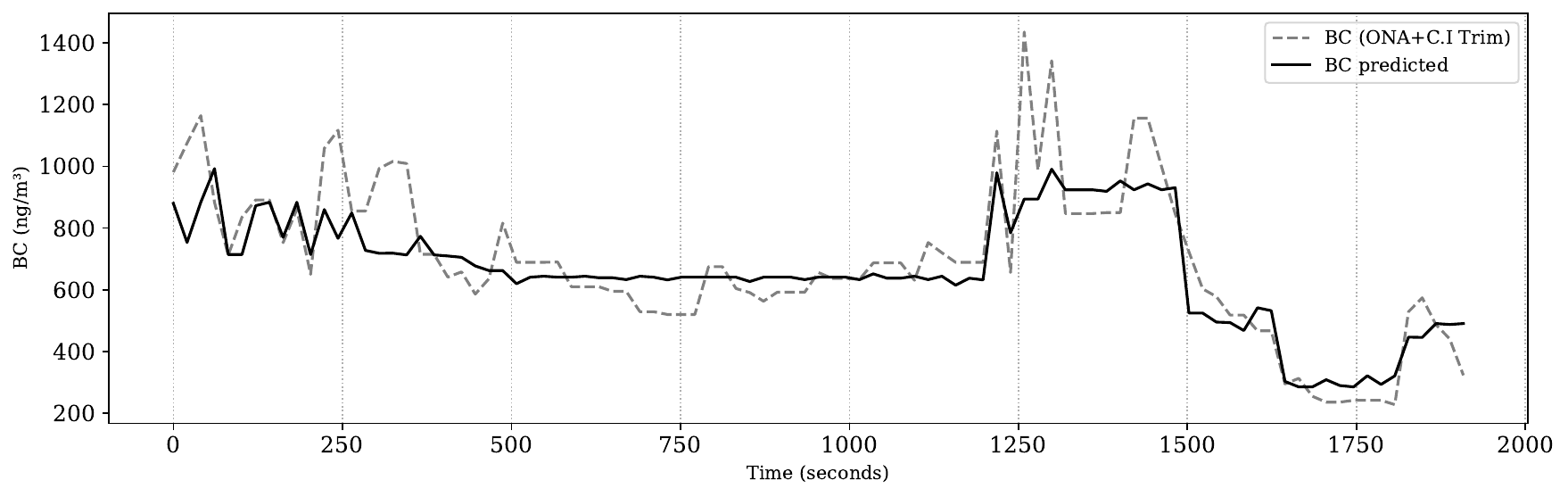}
    \caption{Predicted BC concentration (black) compared against ground truth BC data (dotted) on test set.}
    \label{fig:regressorResultsBC}
\end{figure*}

\subsection{Results}
\subsubsection{Model Selection}
\autoref{tab:resultsAlgo} summarizes the model performance for both our training and testing. We treated Linear Regression as the baseline given its simplicity and effectiveness in providing model interpretability. Where the RMSE value closer to zero and $R^{2}$ closer to 1 is most ideal, our experiments found that the XGBoost regression performs best. The overall regression is statistically significant ($R^{2}$=0.72, p<.001) compared to employing Linear Regression, but it is not significantly different from a Random Forest regressor and Gradient Boosting. \autoref{fig:regressorResultsBC} plots BC ground truth data against model predictions in our test set, with the results deviating from the true value by approximately 129.42\BCUnit. When temporal structure is enforced using windowed stratified split, the $R^{2}$ (0.30, p<.001) decreases as expected due to limited temporal depth in each set. However, it is notable that RMSE and MAE remain within a comparable range (119.93\BCUnit, 91.11). Furthermore, 5-fold cross-validation on the training data yields an 
$R^{2}$ of 0.91, indicating that the model explains a large proportion of variance on unseen data drawn from the same temporal distribution as the training data.

\subsubsection{Noise Reduction}
Since BC readings obtained from aethalometers are often subject to noise, we employed noise reduction strategies as per Section \ref{sec:noiseStrategy}. In removing outliers, the test set remained completely unchanged. \autoref{tab:noiseResult} summarizes the model performance across different approaches. Applying the ONA method is evidently a significant step in improving our model's performance, boosting the $R^{2}$ value from 0.49 to 0.62. Given that our data sets were collected at different traffic junctions, removing outliers locally in each set before combining allows natural traffic variation to be preserved and ensures that the outlier removal process is appropriate for each context. When comparing global and local trimming with baseline, we found that local trimming yielded better results, but these differences are statistically insignificant. Finally, combining ONA with local trimming emerges as the most effective approach, offering the best overall performance at p<.05.

\begin{figure}
    \centering
	\includegraphics[width=0.75\linewidth]{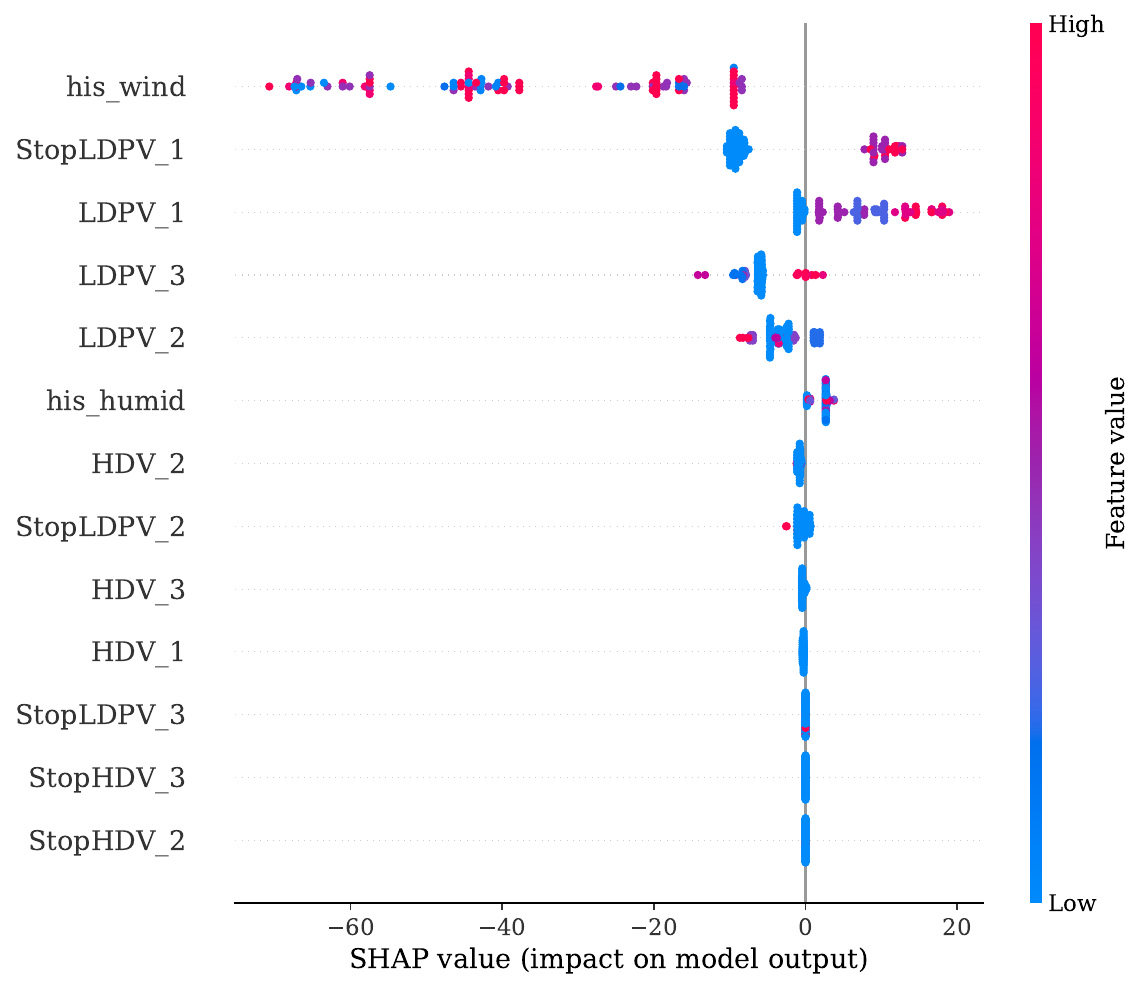}
    \includegraphics[width=0.75\linewidth]{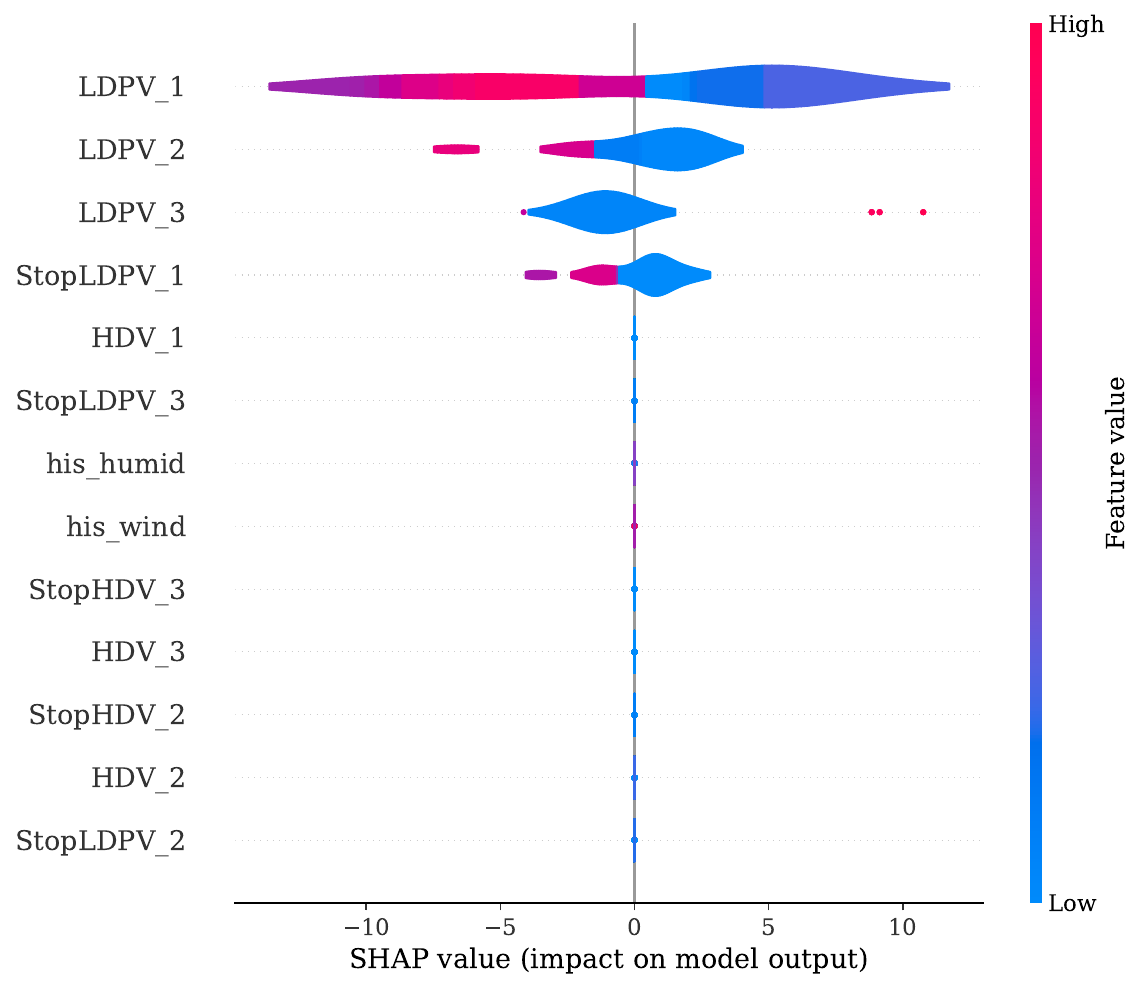}
    \captionof{figure}{SHAP analysis for stratified (top) and windowed stratified (bottom) shows strong similarities.}
    \vspace{1\baselineskip}
    \label{fig:shap}
\end{figure}

\subsection{Features per Case-based Analysis}
\label{sec:caseAnalysis}
We employed SHAP (SHapley Additive ExPlanations) analysis, a game-theoretic approach that assigns each feature a contribution value for individual predictions \cite{lundberg2017unified}. Although the $R^{2}$ values differ substantially between our evaluation schemes, both approaches identify a similar set of features as the strongest contributors, as per \autoref{fig:shap}. The dominant predictors reflect stable relationships between vehicle-related features and BC concentration that do not depend strongly on how data was partitioned.

\begin{figure}
    \centering
    \includegraphics[width=0.8\linewidth]{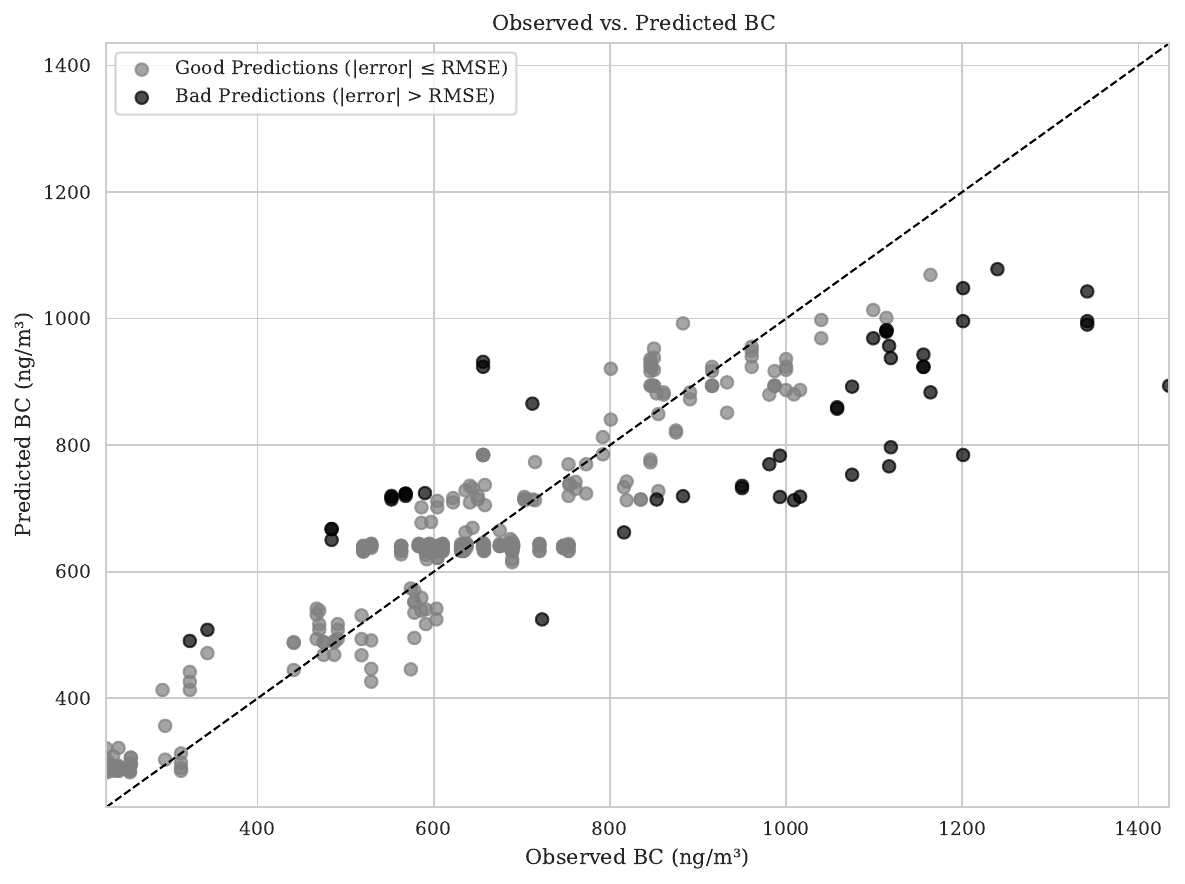}
    \captionof{figure}{Prediction outcomes with smaller errors are defined as those with an error lower than the mean RMSE. Black points represent the worst-case errors, with increasing trend in high-BC scenarios.}
    \label{fig:observed_predicted_bc}
\end{figure}

To better understand the inner workings of the model, we visualize the prediction outcomes in \autoref{fig:observed_predicted_bc}. In this plot, gray points represent predictions with smaller errors (those with a prediction error below the mean RMSE), while black points highlight the worst-case predictions, where the error exceeds the mean RMSE. Notably, the plot illustrates a clear trend of our model performance deteriorating at higher BC concentrations. This result suggests an increased difficulty in prediction in high-BC scenarios, which we conjecture to be influenced by environmental factors. 

\subsubsection{Impact of Environmental Factors}
\begin{figure}
    \centering
	\includegraphics[width=\linewidth]{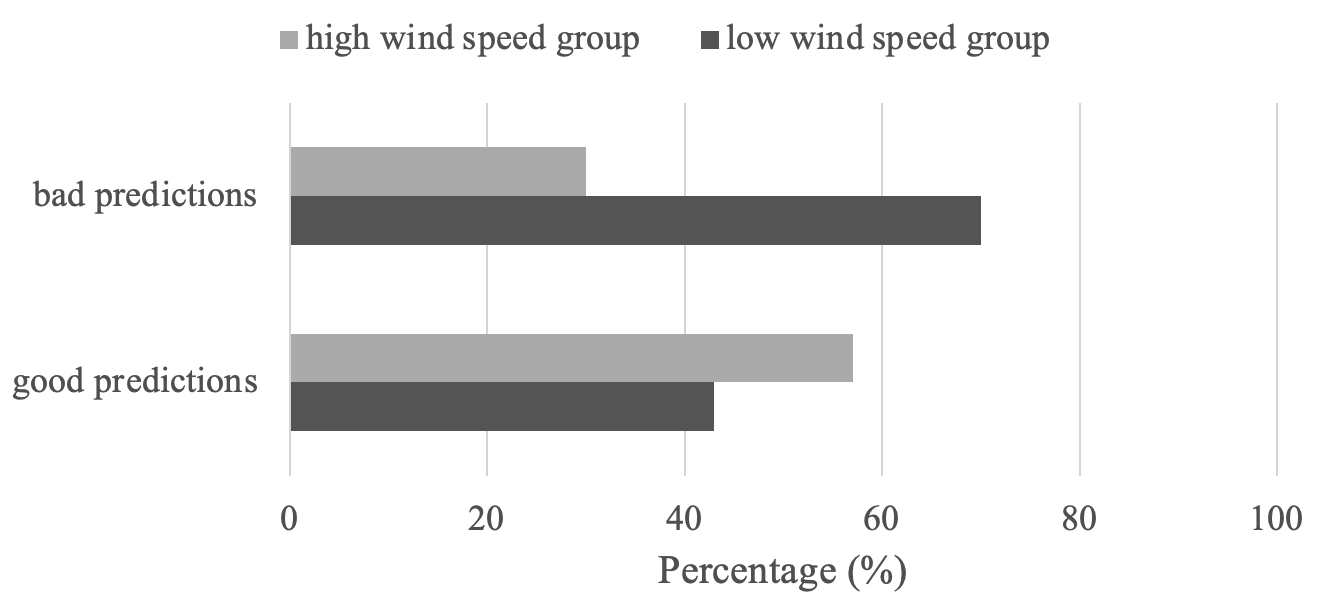}
    \captionof{figure}{Distribution of predictions shows bad predictions attributed to low wind speed than high.}
    \label{fig:BCwindOnly}
\end{figure}

As per \autoref{fig:shap} (top), wind speed (\emph{his\_wind}) consistently emerged as the strongest predictor. This aligns with previous studies, which have repeatedly suggested that moderate wind speeds facilitate pollutant dispersion while lower wind speeds influence BC deposition \cite{weichenthal2014characterizing,hilker2019traffic}. In further examining our predicted outcomes, we found larger prediction errors associated with conditions of low wind speed (70\%), as per \autoref{fig:BCwindOnly}. The difference between good and bad predictions at low wind speed is statistically significant (p<.001). 

\begin{figure}
    \centering
	\includegraphics[width=\linewidth]{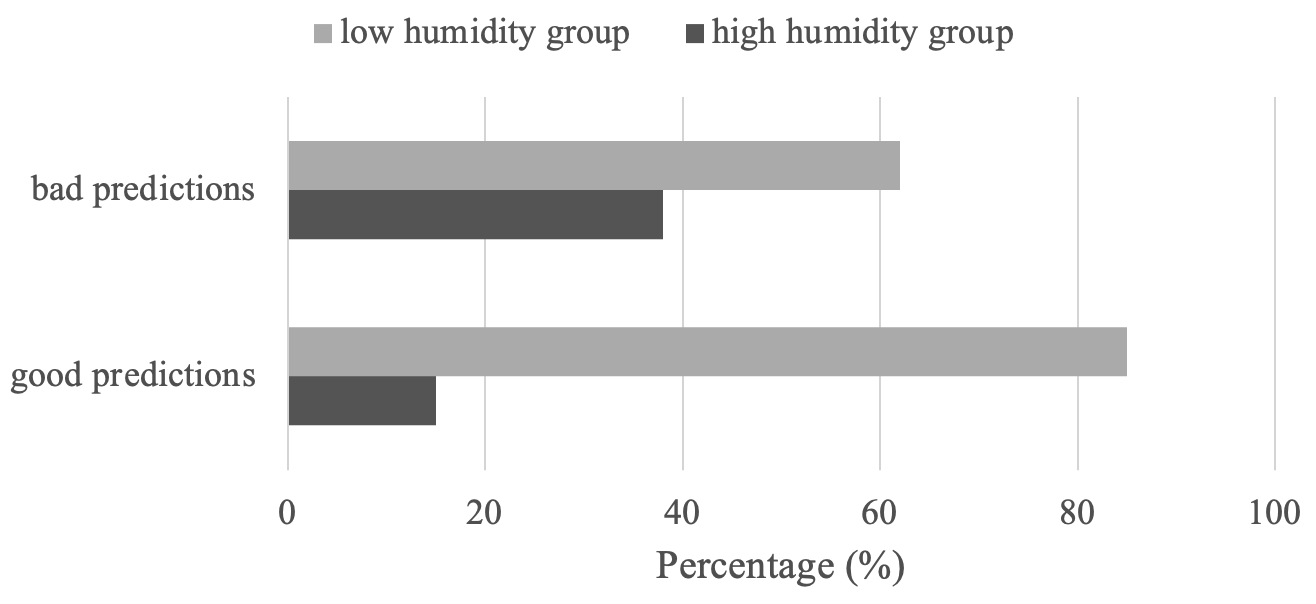}
    \captionof{figure}{Distribution of predictions shows more bad predictions attributed to high humidity than low.}
    \label{fig:BChumidOnly}
\end{figure}

\begin{figure*}[h!]
    \centering
	\includegraphics[width=.75\linewidth]{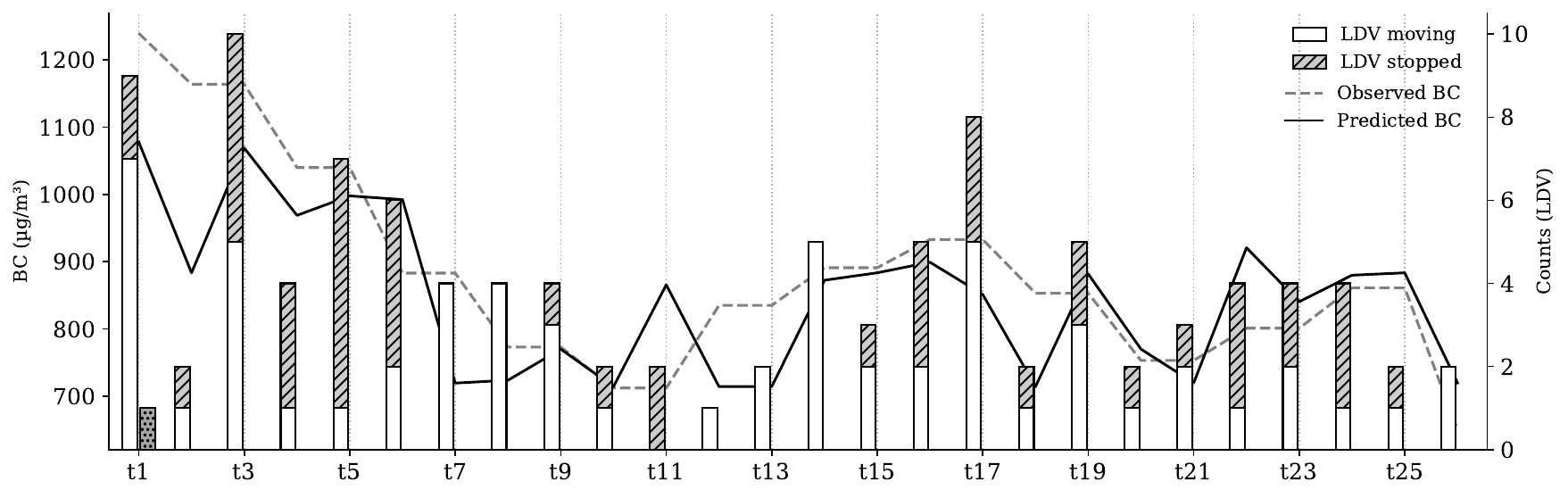}
    \captionof{figure}{Predicted and observed BC concentrations, combined with counts of moving and stopped LDPVs and HDVs over 13 minutes. Predicted values follow the general trend of surges in stopped and moving traffic.}
    \label{fig:resultBest}
\end{figure*}

\begin{figure*}
	\centering
	\includegraphics[width=.75\linewidth]{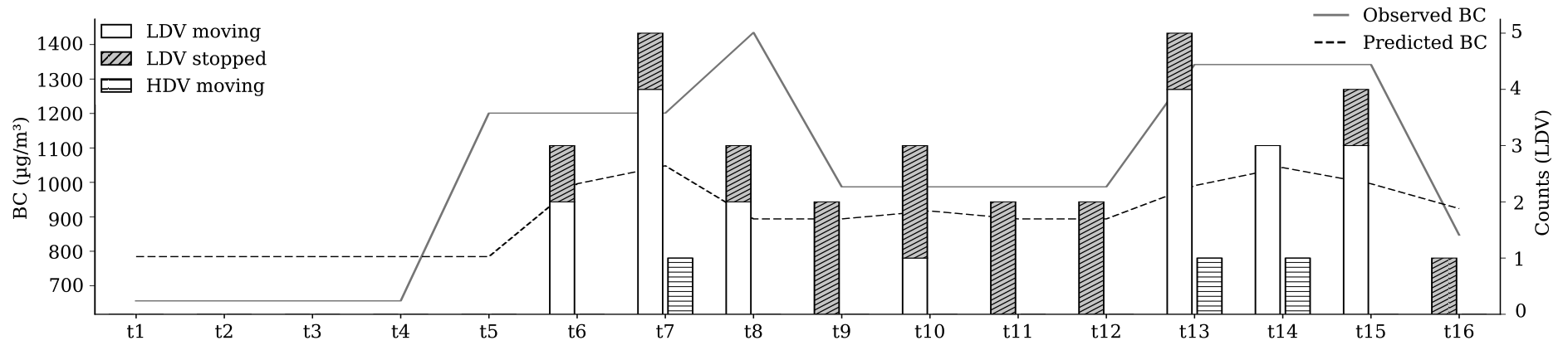}
    \captionof{figure}{Predicted and observed BC concentrations, combined with counts of vehicles. Predicted values do not follow the general trend of observed BC concentration, but aligns with vehicle presence.}
    \label{fig:resultWorst}
\end{figure*}

\begin{figure} [h!]
    \centering
    \includegraphics[width=\linewidth]{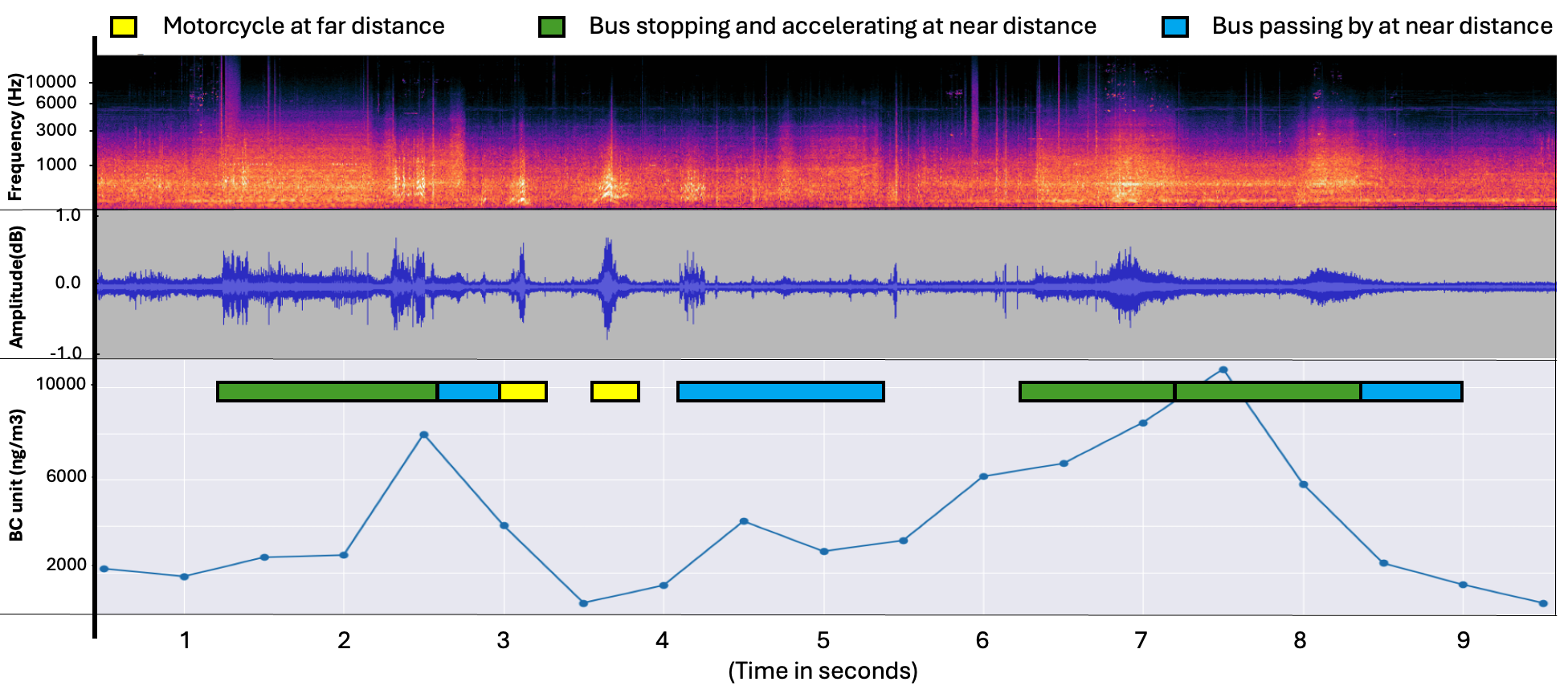}
    \caption{Spectrogram (top) and waveform (middle) show audio differences across bus states; bottom panel shows corresponding BC levels.}
    \label{fig:BCaudio}
\end{figure}

Humidity (i.e., \emph{his\_humid}) is less influential compared to other vehicle emission factor covariates that contribute more significantly to the variation in BC concentrations. As per \autoref{fig:BChumidOnly}, both good and bad predictions were generally associated with low humidity. However, there were more bad predictions associated with high humidity than good ones. This difference is statistically significant at p<.005. Both low wind speeds and high humidity can significantly influence BC concentrations in the air. Under low wind speeds, the atmosphere is less ventilated, meaning that pollutants such as BC are less likely to disperse and can accumulate near the ground \cite{hilker2019traffic}. High humidity can also affect BC concentrations by influencing their physical properties and affecting how they interact with atmospheric processes such as deposition. The combination of these factors creates complex environmental conditions in which predicting BC levels becomes increasingly challenging.

\subsubsection{Impact of Vehicle-based Covariates: Best Case}
With the strongest features associated with the acceleration of passenger vehicles after a full stop and their being in the lane of the road closest to our monitoring setup (i.e., $StopLDPV_{1}$), we visualized the predicted BC values against the ground truth, overlaying corresponding vehicle counts for both moving and stopped LDPV and HDVs in all lanes, as per \autoref{fig:resultBest}. Our observations support the claim that stopped vehicles (and, as a result, their acceleration) have a disproportionately large effect on BC concentration \cite{dons2013street,zhang2019black}. In scenarios where stopped vehicles were present among moving LDPVs, predicted BC concentrations were consistently higher compared to periods of continuous, free-flowing traffic. This result underscores the increased influence of idle or stop-and-go traffic near the monitoring location, likely due to the accumulation of localized emissions and the reduced dispersion under stationary conditions. For example, the highest peak in BC concentration at $t_{3}$ corresponds to localized emission surges of 5 stopped and 10 moving LDPVs. %Larger prediction errors were observed at specific time points ($t_{12}$, $t_{13}$, and $t_{18}$), which coincided with stagnant atmospheric conditions characterized by low wind speed (15 km/h) and high humidity (70\%). As discussed previously, these discrepancies likely reflect the model’s limited ability to capture the effects of reduced dispersion under such conditions.

\subsubsection{Impact of Vehicle-based Covariates: Worst Case} \autoref{fig:resultWorst} charts our worst-case scenario in which the prediction error exceeds the average margin of ±245\BCUnit. We identified two sources of inaccuracy from the raw data logs. The first limitation arises from the small amount of training data available for HDVs. As a result, the model produced errors at $t_{7}$, $t_{13}$, and $t_{14}$, where higher BC concentrations were observed due to the passage of HDVs that were not adequately represented in training. Consistent with this, the SHAP analysis ranked HDV-related characteristics less important, despite strong evidence that such vehicles emit substantially more BC than LDPVs \cite{zheng2017characteristics,zhang2019black}. The second limitation stems from object occlusion in our video data. Along this roadway, a streetcar operates in mixed traffic. Although the streetcar itself is electric and does not emit BC, its presence forces surrounding vehicles to stop and then accelerate once it is safe to move again. These stop–start conditions generate substantial amounts of BC. However, our technique was unable to fully capture them because the streetcar occluded many of the stopped vehicles and led to undercounting of vehicle types and misclassification of stopping and acceleration events. 

After excluding the affected data points, model performance improved, with RMSE decreasing to 114 \BCUnit. Ground-level video surveillance often suffers from object occlusion, and this effect would be mitigated in datasets captured from aerial or elevated camera angles.\newline

\noindent \textbf{Key Takeaway:} Our BC prediction model integrates emission-related covariates derived from traffic video recordings with environmental variables extracted from daily weather data. Using an XGBoost regression framework, the model achieved an $R^{2}$ of 0.72 and an RMSE of 129.42 (\BCUnit). This result indicates that 72\% of the variance in BC levels can be explained by the selected features. The corresponding MAE increased only modestly from 71 (train) to 91 (test), suggesting limited overfitting and that the model generalizes reasonably well to unseen data within the available dataset. Patterns of stop-and-accelerate behavior, particularly for vehicles in the lane closest to the monitoring setup, align with established findings in the literature. What is novel here is that these patterns are derived from computer vision features, demonstrating the potential of vision-based approaches to capture real-world emission dynamics.
\section{Discussion}
\label{sec:discussion}
We proposed a CV-based method to estimate the amounts of BC concentration in traffic using traffic video recordings as a primary data source. Here, we discuss the limitations and implications of our findings.

\subsection{Improving Model Performance}

This study focuses on the use of CV techniques to extract standard vehicle characteristics. However, our model implementation has limitations, particularly in training one that differentiates green vehicles from conventional vehicles. Other techniques can be extended. For instance, license plate detection can reveal codes of fuel type, as in Ontario, Canada, where plates beginning with “GV” are designated for green vehicles. In practice, this implementation means extracting only the minimal information required, while avoiding the storage of full license plate data or allowing the data to be repurposed for other uses.

Beyond vision-based cues, acoustic sensing holds promise for distinguishing engine types, and our work continues to examine how audio signals \cite{dawton2020initial} can be integrated. We are currently extending audio-sensing capabilities, as per prior work \cite{murovec2024zero,wieczorkowska2018spectral}, extracting features such as zero crossing rate, root mean square energy, spectral roll-off, and spectral centroid. As illustrated in \autoref{fig:BCaudio}, we observed clear peaks in BC concentration shortly after the presence of buses near our audio-sensing prototype. 

We anticipate continuous audio monitoring along congested highways to be overwhelming and largely impractical due to background noise and traffic volume, but targeted deployment in community zones can still provide meaningful value.

\subsection{Monitoring for Environmental Justice}
Environmental justice research on BC and ultrafine particles (UFPs) remains relatively limited in general, but urban research in Canada and the US has highlighted emerging concerns \cite{batisse2025examining,chambliss2021local}. Specifically, as an example, Elford \emph{et al.} reported that areas in Toronto with a higher prevalence of immigrants faced greater exposure to UFPs during walks to school \cite{elford2021associations}, supporting broader findings from U.S. cities where low-income and visible minority groups, particularly Black, Hispanic and Latino populations, are disproportionately exposed to high BC concentrations \cite{chambliss2021local}. Our study is geographically limited to several streets in the Downtown Toronto region, where traffic is predominantly dense. While the current dataset does not capture seasonal variations, it does include variation in wind speed, which is highly relevant for BC prediction. Our experiments demonstrated the technical feasibility through strong correlations between video-derived vehicle inferences and BC concentration. At the same time, these results provided us with clearer guidance in conducting a long-term regional study, including additional sensing tools that we will now consider to improve prediction. 

Unlike regulatory weather stations or low-cost air quality sensors, which are often more prevalent in predominantly white and higher income neighborhoods \cite{kelp2023data}, traffic surveillance cameras tend to be more available in urban environments. This availability makes traffic surveillance cameras a useful data source for near-source BC estimation from traffic. While the disproportionate placement of traffic surveillance cameras in lower-income communities has raised equity concerns \cite{VisionZeroNetwork2025}, this existing infrastructure could be leveraged for positive purposes. Instead of reinforcing punitive or enforcement-heavy practices, traffic cameras can support environmental and public health monitoring by providing insight into traffic pollution and guiding interventions that benefit the very communities most affected. Furthermore, BC concentrations are influenced by broader urban and regional environmental factors, including local topography and contributions from other sources such as wildfires, industrial emissions, and residential heating. Hence, weather stations equipped with microaethalometers remain the gold standard for capturing complex regional and atmospheric dynamics. Instead, our system focuses on near-source estimations of street-level traffic, complementing research-grade instruments.

From a policy perspective, having more localized BC estimates plays a critical role in strengthening the evidence base for environmental justice efforts. Beyond identifying areas of concern, such data can directly inform targeted interventions, including reducing vehicle idle near traffic hotspots and school zones, redirecting HDVs away from residential neighborhoods during peak pedestrian hours, and optimizing traffic signal timing in areas with frequent peaks of stopped vehicles \cite{zalzal2022fifteen}.
\section{Conclusion}
Traffic video recordings are valuable resource for transportation planning, public safety, and urban development. While ML-driven applications in this domain have largely focused on improving driver behavior and enabling dynamic traffic management, their potential to study the environmental impacts of traffic remains underexplored. This gap is especially critical given the urgent need to address pollution in urban environments, where traffic is a major source of BC emissions. However, the subject of BC is particularly challenging to study due to the high cost and complexity of instrumentation and gathering measurements compared to other pollutants. We proposed an ML-driven approach that determines vehicle emission factor covariates to estimate BC concentration. Our XGBoost regression model achieved an $R^{2}$ of 0.72 and an RMSE of 129.42 \BCUnit. Modeling BC concentration prediction is inherently constrained from reaching an $R^{2}$ close to 1 because these pollutants are influenced not only by this source but also by regional factors (e.g., in nearby towns), urban structures, and other local sources (e.g., residential exhaust). Since our approach focuses only on vehicle-related local factors, an $R^{2}$ of 0.72 is a promising outcome, reflecting the natural but partial scope of predictors. The most influential predictors identified were wind speed, vehicle acceleration after stopping, and the distance between the vehicle and the measurement device. These findings align with prior research showing that wind speed affects pollutant dispersion, with BC concentrations highest near sources and decreasing with distance. Our work continues with a larger data collection campaign spanning the Greater Toronto Area across seasonal changes, expanding video coverage and incorporating audio.

\begin{acks}
This work is supported by the Natural Sciences and Engineering Research Council of Canada. [RGPIN-2025-06914]
\end{acks}

\bibliographystyle{ACM-Reference-Format}
\bibliography{references}

\newpage
\onecolumn  % switch to single-column for appendix

\begin{appendices}
% =========================
% APPENDIX
% =========================
\section{Dataset Excerpt}
This appendix describes the raw and processed dataset we used for the experiments in this paper.

\begin{figure}[ht]
    \centering
    \begin{lstlisting}[style=thin,label={lst:ae51}]
    Date,Time,Ref,Sen,ATN,Flow,Pcb temp,Status,Battery,BC,Ona_#_pts_avg
    2024/11/04,18:49:00,890665,921559,-3.40984263756,100,19,0,98,,NULL
    2024/11/04,18:49:30,890783,921490,-3.3891073947402,99,19,0,98,2379,3
    2024/11/04,18:50:00,890907,921527,-3.3792031816136,99,19,0,98,1136,3
    2024/11/04,18:50:30,890941,921473,-3.369526908093,100,19,0,98,1099,3
    2024/11/04,18:51:00,891037,921486,-3.3601631390269,100,19,0,98,1064,2
    \end{lstlisting}
    \caption{Raw data samples collected using the microaethalometer, microAeth AE51, collected at 30 seconds interval in .CSV}
\end{figure}

\begin{figure}[ht]
    \centering
    \begin{lstlisting}[style=thin,language=bash,label={lst:eventlog}]
     car_line1 : 2771 2024-11-05 17:23:02
     car_line1 : 2775 2024-11-05 17:23:03
     car_line2 : 2790 2024-11-05 17:23:05
     truck_line2 : 2777 2024-11-05 17:23:07
     person_line1 : 2785 2024-11-05 17:23:09
     bicycle_line1 : 2795 2024-11-05 17:23:09
    \end{lstlisting}
    \caption{Features of vehicles in traffic extracted from the video processing pipeline. When an object is first seen, the system assigns a unique ID (e.g., 2771) and records their respective lane and timestamp.}
\end{figure}

\begin{figure}[ht]
    \centering
    \begin{lstlisting}[style=thin,label={lst:dataset-row}]
    {
      "Time": "2024-11-05 17:34:50",
      "BC": 1202,
      "BC post": 1114,
      "car_line1": 1,
      "car_line3": 3,
      "truck_line2": 1,
      "car_line2": 2,
      "car_line1_stop": 0,
      "truck_line2_stop": 0,
      "truck_line3": 0,
      "truck_line1": 0,
      "traffic": 0.65,
      "history_temperature": 19.1,
      "history_wind_speed": 24.5,
      "history_humidity": 67,
      "forecast_temperature": 18.7,
      "forecast_wind_speed": 9.9,
      "forecast_humidity": 68
    }
    \end{lstlisting}
    \caption{This is an example row of what our final dataset looks like. The variable ``BC post'' indicates BC after applying Optimized Noise-Reduction Algorithm (ONA).}
\end{figure}

\section{Hyperparameter Tuning}
We tuned model hyperparameters using cross-validated grid search in \texttt{scikit-learn}. Each model was wrapped in a common \texttt{Pipeline} so that parameters are referenced with the double-underscore convention, for example \texttt{model\_\_max\_depth}. The search spaces mirror the configuration used in our code.

\begin{table}[h!]

\centering
\small
\begin{tabularx}{\linewidth}{@{}>{\raggedright\arraybackslash}l
                               >{\raggedright\arraybackslash}l
                               Y@{}}
\toprule
\textbf{Model} & \textbf{Parameter} & \textbf{Values} \\
\midrule
LinearRegression & (none) & No hyperparameters tuned \\[2pt]

\multirow{3}{*}{RandomForestRegressor}
 & \texttt{model\_\_min\_samples\_split} & [2, 5] \\
 & \texttt{model\_\_n\_estimators}       & [30, 50, 100, 200] \\
 & \texttt{model\_\_max\_depth}          & [None, 5, 6, 7, 9, 10] \\[2pt]

\multirow{3}{*}{GradientBoostingRegressor}
 & \texttt{model\_\_n\_estimators}       & [20, 50, 100, 200, 250] \\
 & \texttt{model\_\_learning\_rate}      & [0.01, 0.05, 0.1, 0.2] \\
 & \texttt{model\_\_max\_depth}          & [3, 5, 6, 7] \\[2pt]

\multirow{2}{*}{SVR}
 & \texttt{model\_\_C}                   & [0.1, 1, 10] \\
 & \texttt{model\_\_epsilon}             & [0.01, 0.1, 1] \\[2pt]

\multirow{3}{*}{XGBRegressor}
 & \texttt{model\_\_n\_estimators}       & [20, 50, 100, 200, 250] \\
 & \texttt{model\_\_learning\_rate}      & [0.01, 0.05, 0.1, 0.2] \\
 & \texttt{model\_\_max\_depth}          & [3, 5, 6, 7] \\
\bottomrule
\end{tabularx}
\end{table}

\noindent We ran a cross-validated grid search per model and selected the configuration with the best mean validation score. Unless noted otherwise, we used five-fold cross-validation, refit the best configuration on the full training split, and reported scores on the held-out test split. A fixed \texttt{random\_state} ensured repeatability.
\begin{itemize}
  \item \textbf{RandomForestRegressor} tunes tree count (\texttt{n\_estimators}), depth (\texttt{max\_depth}), and split size (\texttt{min\_samples\_split}) to control ensemble capacity and variance.
  \item \textbf{GradientBoostingRegressor} balances \texttt{n\_estimators} and \texttt{learning\_rate}; \texttt{max\_depth} limits individual tree depth.
  \item \textbf{SVR} primarily depends on \texttt{C} (regularization) and \texttt{epsilon} (tube width). We used the default RBF kernel unless specified.
  \item \textbf{XGBRegressor} varies \texttt{n\_estimators}, \texttt{learning\_rate}, and \texttt{max\_depth} under the squared-error objective.
\end{itemize}

\begin{figure}[h!]
    \centering
    \begin{lstlisting}[style=thin,language=Python]
    search = GridSearchCV(
        estimator=pipe,                 # e.g., Pipeline([('prep', ...), ('model', ...)])
        param_grid=models[name]['params'],
        scoring='neg_root_mean_squared_error',
        cv=5, n_jobs=-1, refit=True
    )
    search.fit(X_train, y_train)
    best_params = search.best_params_
    cv_table   = search.cv_results_
    \end{lstlisting}
    \caption{Skeleton used for cross-validated grid search}
    \label{fig:placeholder}
\end{figure}

\section{Database Schema for Dashboard}

We use Docker Compose to start the dashboard and its services from a Compose file named \texttt{infrastructure.yaml}. The commands below assume you run them at the repository root.

\begin{lstlisting}[style=thin,language=bash]
# Bring services up (detached). 
docker compose -f infrastructure.yaml up -d
\end{lstlisting}

\begin{lstlisting}[style=thin,language=bash]
# Check status. 
docker compose -f infrastructure.yaml ps
\end{lstlisting}

\begin{lstlisting}[style=thin,language=bash]
# Stop and remove the stack.
docker compose -f infrastructure.yaml down
\end{lstlisting}

\begin{figure}[h!]
  \centering
  % Replace the filename below with your actual exported diagram (PDF/PNG)
  \includegraphics[width=0.8\linewidth]{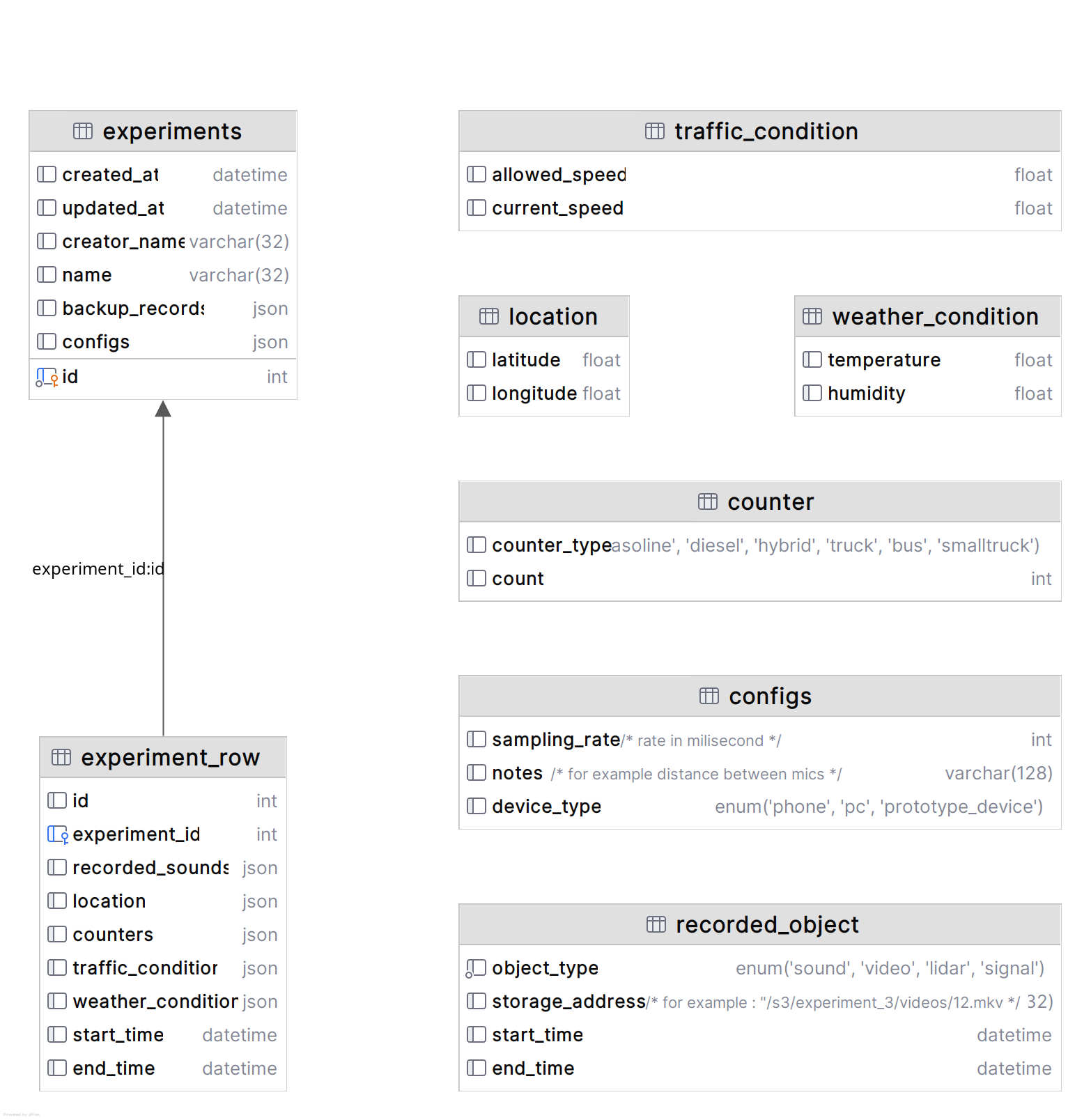}
  \caption{Database schema for our data collection dashboard.}
  \label{fig:early-db-schema}
\end{figure}

\begin{figure}[tb!]
    \centering
	\includegraphics[width=\linewidth]{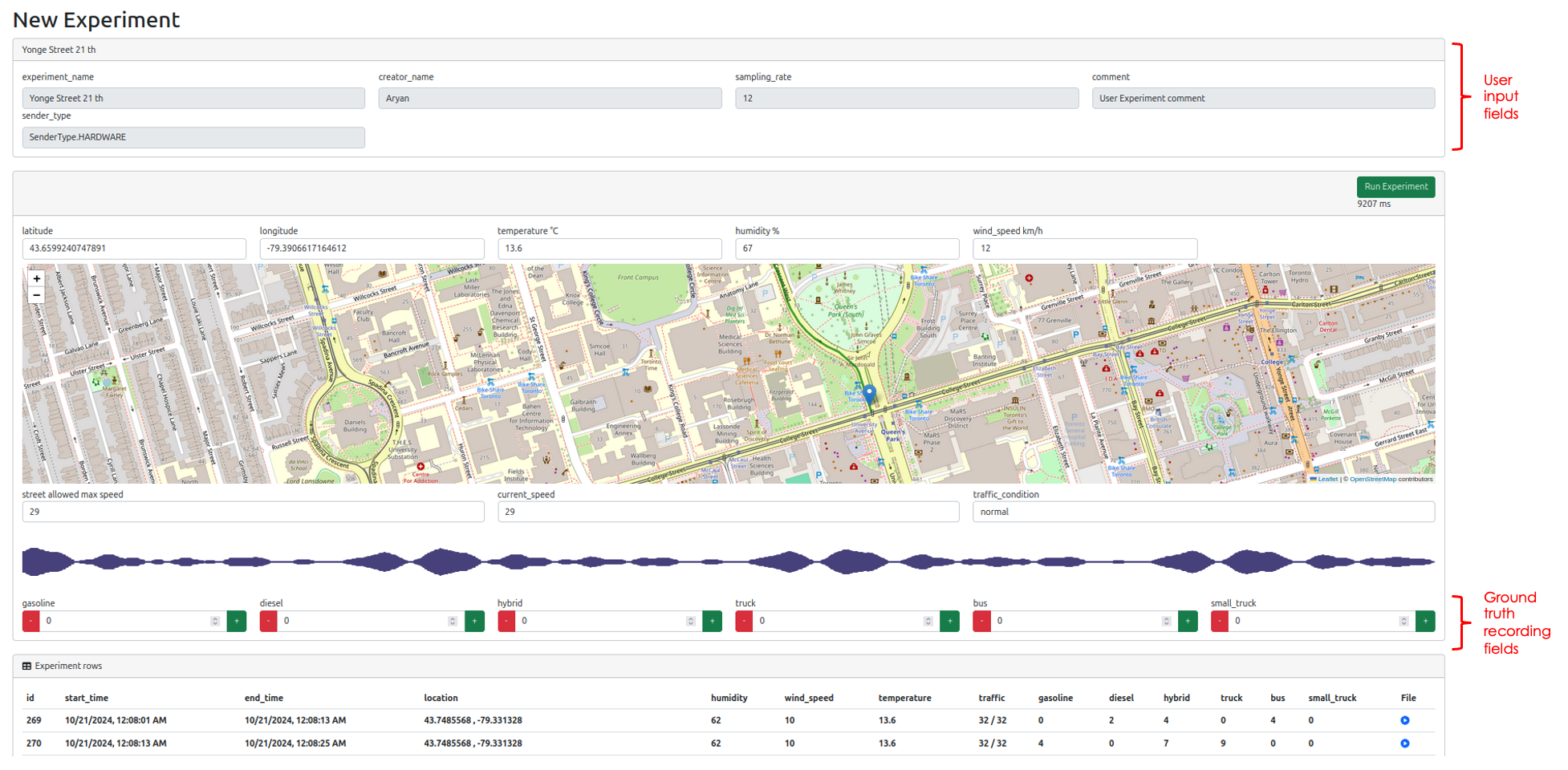}
    \captionof{figure}{Online dashboard consolidating data streams and manual observations into a structured format.}
    \label{fig:dashboard}
\end{figure}
\end{appendices}

\end{document}